\documentclass[10pt,twocolumn,letterpaper]{article}

\usepackage[pagenumbers]{cvpr} 

\usepackage{multirow}
\usepackage{array}
\newcolumntype{C}[1]{>{\centering\arraybackslash}p{#1}}

\usepackage{listings} 
\usepackage[most]{tcolorbox}

\newtcbinputlisting{\styledfileinput}[3][text]{%
  listing file={#2}, listing only,
  breakable, enhanced jigsaw,
  colback=gray!10, colframe=gray!10,
  boxrule=0pt, arc=0pt,
  left=4pt, right=4pt, top=2pt, bottom=2pt,
  listing options={language=#1,#3},
}
\definecolor{cvprblue}{rgb}{0.21,0.49,0.74}
\usepackage[pagebackref,breaklinks,colorlinks,allcolors=cvprblue]{hyperref}

\def\confName{CVPR}
\def\confYear{2027}

\title{Scene Retargeting: Learning Object Placement with Analogical Transfer \vspace{-0.5em}}

\author{
Minkwan Kim$^{1}$ \quad
Junho Kim$^{1}$ \quad
Seungmin Lee$^{1}$ \quad
Changwoon Choi$^{1}$ \quad
Young Min Kim$^{1,2}$ {\small \phantom{ }}\\
{\small $^{1}$ Dept. of Electrical and Computer Engineering, Seoul National University} \\
{\small $^{2}$ Interdisciplinary Program in Artificial Intelligence and INMC, Seoul National University}
}

\begin{document}
\twocolumn[{
\renewcommand\twocolumn[1][]{#1}
\maketitle
\begin{center}
    \centering
    \captionsetup{type=figure}
    \vspace{-0.5em}
    \includegraphics[trim={0, 0, 0, 0}, clip, width=\linewidth]{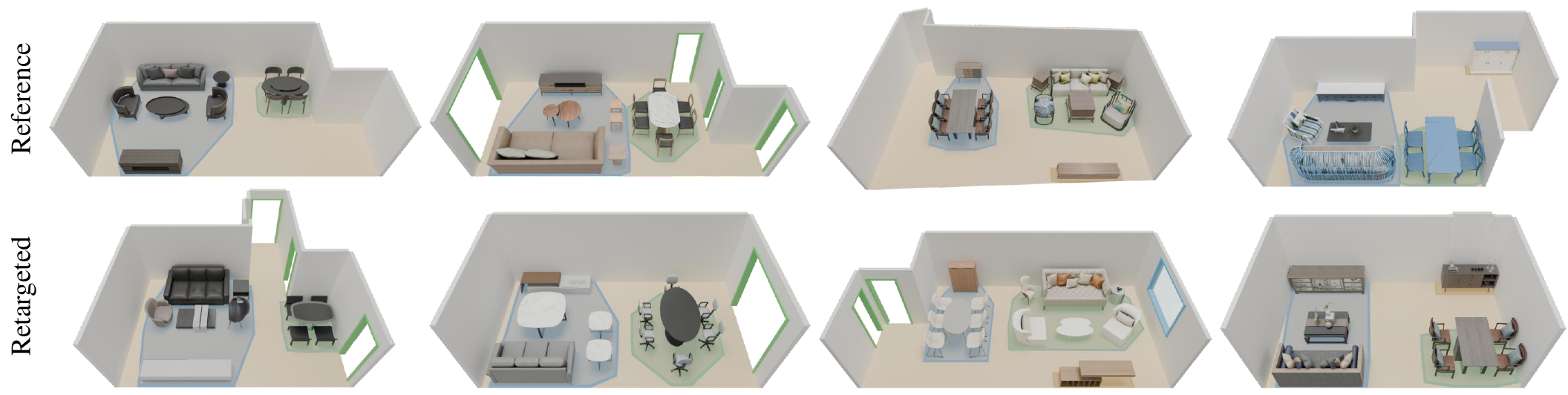}
    \vspace{-1.0em}
    \captionof{figure}{\textbf{Scene Retargeting.} 
    Given a reference scene, our method transfers its functional arrangement to target rooms with different floor plans and object inventories, preserving the arrangement within each functional group under physical constraints.
    }
    \label{fig:teaser}
\end{center}
}]
\maketitle
\begin{abstract}
Interactive simulations of embodied AI or spatial computing applications build on realistic 3D scenes that support daily activities.
However, sparse, irregular layout structures impose scene-specific physical constraints, making it hard to define a generalizable framework for generating similar functional context.
We formalize \textbf{Scene Retargeting} as transferring the functional arrangement of a reference room to a new floor plan and object inventory, rather than relying on textual descriptions or standardized pairwise relationships.
Our cluster-wise transfer reproduces the arrangement within each functional group under mismatched inventories and adapts group placement to the target floor plan.
We optimize to preserve the rich semantic context of individual clusters by respecting the spatial distribution of foundation features.
We can then impose physical constraints to refine wall contacts, pairwise alignment, or clear passageways and openings.
Our framework outperforms state-of-the-art methods on layout generation on the 3D-FRONT dataset, and demonstrates downstream applications including real-to-sim transfer, analogical trajectory transfer, and multi-reference composition.
\end{abstract}    
\section{Introduction}
\label{sec:intro}
Synthesizing plausible 3D scene layouts provides the foundation for diverse daily interactions and can serve as a critical path to deploy embodied agents or AR/VR applications~\cite{ProcTHOR, SAGE, DigitalCousines}.
Manually designing realistic layouts is labor-intensive and hard to scale, necessitating automated synthesis.
Previous works adapt generative models to regress 2D rotations and translations of objects within the given floorplan~\cite{Lego-net, Diffuscene, Midiffusion}.
However, unlike text, images, or videos, data-driven generation is not trivial under sparse, irregular structural restrictions.
A synthesized layout must faithfully reproduce the semantic context of \textit{scene function} for common daily interactions, while maintaining \textit{geometric plausibility} under scene-specific constraints.
Scene function is implicit and nuanced, while complex geometric relationships are defined for single objects, pairs, or nearby groups, with a mixture of rigid and flexible measures.

We propose \textbf{Scene Retargeting}, which transfers the functional arrangement of a reference scene to a new room while respecting its geometric constraints.
Given a reference scene, scene retargeting generates a target scene of similar function by arranging a target object inventory in its floor plan, as illustrated in~\cref{fig:teaser}.
Previous approaches to 3D layout generation account for contextual constraints with LLM/VLMs~\cite{Layoutgpt, I-design, Holodeck, Holodeck2.0} or scene graphs~\cite{Instructscene, Echoscene, Commonscenes}. 
Such abstractions coarsely capture symbolic descriptions or predefined pairwise relationships, which are insufficient to account for the functional context of diverse interactions relative to multiple objects or to express the intricate shape variations of individual furniture and the layout of walls, windows, and doors.
Scene retargeting draws the arrangement of each functional group from a single exemplar and aims for faithful transfer under the constraints imposed by the target floor plan and inventory.

We approach scene retargeting as decomposing and re-structuring clusters of exemplar scenes.
Given different object instances and a target floor plan, no correspondence between objects is given in advance, and even once established, replicating the transformation cannot preserve the same semantics in relation to walls or doors. 
We group nearby objects into a small number of clusters within the reference scene using both spatial coordinates and semantic features.
Since the clusters serve as the primary unit of transfer, they reduce the degrees of freedom from a pose per object to a pose per group, and the arrangement of reference objects within each group is carried over as a unit and instantiated with the target inventory.
Cluster-wise transfer thus preserves the arrangement within each group while adapting the initial placement of groups to the target floor plan, even with different numbers of objects and unknown correspondences.

We formulate a three-stage hierarchical framework that preserves the functional context while being adaptive to the target geometry.
We first extract clusters from the reference scene by grouping neighboring objects of similar semantic features, and stage 1 locates the clusters on the target floor plan using a learned placement network.
In stage 2, we place objects within each cluster by attending to 3D features of the transferred reference layout,
which encode the spatial distribution of desired semantic relationships beyond what category labels or language-based descriptions could express.
After the previous stages handle inter-cluster and intra-cluster placements, respectively, stage 3 establishes explicit object correspondences and refines local arrangements to respect pairwise relations or wall contacts, while ensuring physical validity such as clearing passages or avoiding penetrations.

Our key contributions are summarized as follows: 
1) We formalize the task of Scene Retargeting, which transfers the functional arrangement of a single exemplar to new floor plans and object inventories.
2) We propose cluster-level placement on the target floor plan followed by object-level arrangement conditioned on pretrained 3D foundation model features, capturing rich spatial context that language descriptions or scene graphs cannot express.
3) We propose a three-stage framework that preserves the arrangement within each functional cluster while adapting cluster placement to the target room under physical constraints.
The resulting pipeline is applicable to various downstream scenarios including real-to-sim scene transport, analogical trajectory transfer, and multi-reference composition.

\section{Related Work}
\label{sec:related_work}
\subsection{Learning-based 3D Layout Generation} 
Most learning-based indoor layout generation methods model a joint probability distribution over object attributes within a room. 
Early approaches synthesize objects autoregressively, one at a time~\cite{Atiss}. 
Subsequent works predict object attributes simultaneously through denoising frameworks~\cite{Lego-net, Diffuscene, Midiffusion, Physcene}. 
Alternatively, several methods incorporate intermediate scene graphs to explicitly model pairwise relations rather than relying solely on spatial decoders~\cite{Instructscene, Freescene, Echoscene, Commonscenes, Flowscene}.
Recently, LaviGen~\cite{LaviGen} places objects sequentially in native 3D space using an adapted 3D diffusion model~\cite{Trellis}. 
While these paradigms excel at statistically plausible generation, their reliance on coarse inputs like text or floor plans prevents them from faithfully reproducing a specific exemplar layout.
Consequently, variation across generated samples stems from stochastic sampling rather than adherence to a target layout.

\subsection{LLM/VLM-driven Scene Synthesis}
Several recent works adopt natural language as a flexible interface, leveraging the spatial reasoning of LLMs and VLMs~\cite{GPT, Gemini} for intent-driven layout generation. 
One line of work translates 3D scene synthesis into structured text or code, prompting LLMs to perform symbolic spatial reasoning, in-context layout generation, or multi-agent constraint coordination~\cite{Layoutgpt, I-design, Holodeck, Holodeck2.0, SAGE, Artiscene}. 
To better bridge text with continuous 3D space, other approaches leverage VLMs to predict parameterized scene representations, which are subsequently refined via differentiable optimization~\cite{LayoutVLM, Sceneweaver}. 
While these methods offer intuitive user control, language remains a fundamentally ambiguous modality for capturing metric 3D geometry. 
Verbalizing a complex 3D arrangement inevitably discards precise spatial metrics, retaining only coarse relational concepts rather than the exact geometric layout. 
In contrast, our approach bypasses textual abstractions and directly leverages features from pretrained 3D foundation models~\cite{Concerto}, preserving the fine-grained metric relationships required for accurate scene retargeting.
\begin{figure*}[t]
    \centering
    \includegraphics[width=\linewidth]{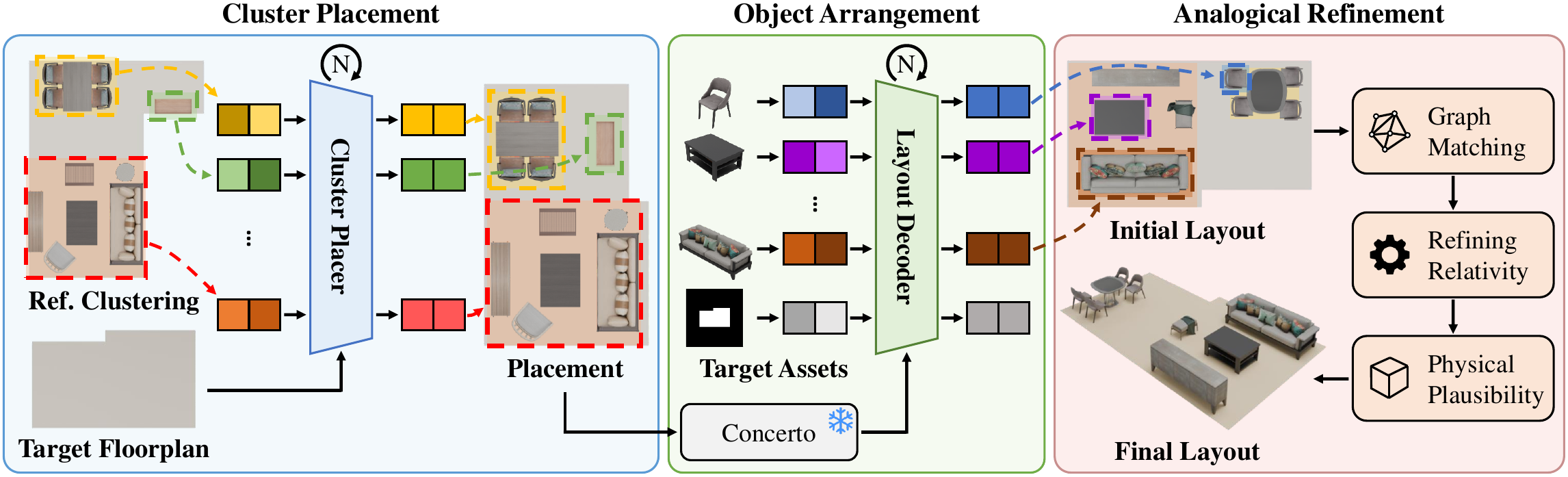}
    \caption{\textbf{Method Overview of Scene Retargeting.} In \textbf{stage 1} (Cluster Placement, blue), functional clusters decomposed from the reference scene are placed onto the target floor plan by the cluster placer. In \textbf{stage 2} (Object Arrangement, green), object tokens cross-attend to spatial context features within the layout decoder to assign object-level poses. \textbf{Stage 3} (Analogical Refinement, red) matches objects across scenes, optimizes pairwise relations and wall clearances, and enforces physical validity to produce the final 3D layout.}
    \label{fig:method_overview}
\end{figure*}
%

\subsection{Exemplar-based Layout Synthesis}
A third line of work refines or rearranges an existing layout rather than sampling a new one from scratch. 
Early example-based methods learn spatial priors from scene collections~\cite{Example-based, MakeitHome}, while optimization frameworks adjust arrangements against design rules or cost functions~\cite{Interactive_Furniture_Layout, MakeitHome}. 
Closest to our formulation is 3D scene analogies, which maps spatial relationships between scenes to transfer trajectories~\cite{SceneAnalogy, ATT}. 
However, such mapping presupposes that both scenes are fully furnished, whereas scene retargeting arranges an empty target room. Our framework differs in two fundamental ways. 
First, the arrangement within each functional group is drawn from a single reference scene rather than a dataset-derived statistical prior, with a learned prior used only to place the groups in the target room.
Second, while existing methods operate within a fixed room geometry, scene retargeting transfers the functional arrangement across distinct room boundaries and mismatched object inventories.
Consequently, our analogical refinement derives its optimization terms directly from the reference scene rather than from design rules or learned costs.
\section{Method}
\label{sec:method}

\subsection{Overview}
\label{subsec:Overview}
\noindent\textbf{Task Formulation.}
We formalize \textbf{Scene Retargeting} as the transfer of the functional arrangement of a reference scene $\mathcal{S}_R = \{(o_i^R, \mathbf{p}_i^R)\}_{i=1}^{N_R}$ to an empty target room. 
Here, $o_i^R$ denotes 3D asset geometry and $\mathbf{p}_i^R = (x_i^R, y_i^R, \theta_i^R) \in \mathbb{R}^2 \times [-\pi, \pi)$ represents its 2D floor-plane pose.
The rooms are defined by a floor plan polygon $\mathcal{F}_R, \mathcal{F}_T \subset \mathbb{R}^2$ whose edges form the room's walls and architectural openings $\mathcal{B}_T = \{b_m\}_{m=1}^{M}$, where each $b_m$ specifies a door or window along the boundary.
Each opening induces a walk-through clearance zone that must remain unobstructed for the room to stay navigable.
Given an unarranged target inventory $\mathcal{O}_T = \{o_j^T\}_{j=1}^{N_T}$, our objective is to predict target poses $\mathbf{P}_T = \{\mathbf{p}_j^T\}_{j=1}^{N_T}$ placing each asset $o_j^T$ onto $\mathcal{F}_T$.
The resulting layout $\mathcal{S}_T=\{(o_j^T, \mathbf{p}_j^T)\}_{j=1}^{N_T}$ must preserve the arrangement within each cluster of $\mathcal{S}_R$ while adapting cluster placement to $\mathcal{F}_T$ and $\mathcal{B}_T$ under physical plausibility.
 
\noindent\textbf{Pipeline Overview.}
Our framework comprises three stages, illustrated in~\cref{fig:method_overview}. 
Given different rooms and object inventories, we cannot directly transfer absolute coordinates.
Instead, we transfer the scene by introducing \textit{clusters} as an additional spatial granularity.
%
Stage 1 (\cref{subsec:cluster_transfer}) partitions $\mathcal{S}_R$ into functional clusters, which serve as the primary units of transfer, and places them onto $\mathcal{F}_T$ with a learned placement network. 
Stage 2 (\cref{subsec:object_placement}) resolves object-level poses within each placed cluster by attending to the transferred spatial context. 
Stage 3 (\cref{subsec:analogical_refinement}) is parameter-free: it establishes reference–target correspondences and refines target poses $\mathcal{S}_T$ against pairwise relations measured on the reference $\mathcal{S}_R$ while enforcing physical validity, alignment against wall $\mathcal{F}_T$, and clearances of opening $\mathcal{B}_T$.

\subsection{Cluster-Level Transfer}
\label{subsec:cluster_transfer}
Placing objects individually within a new boundary lacks a mechanism to preserve related assets together, so the relationship and semantics of the exemplar may be lost.
To address this, we group co-located and semantically associated objects into \textit{functional clusters} and take these as the primary unit of transfer.
The arrangement of reference objects within each cluster is then carried over as a unit, leaving the placement network to predict only where each cluster belongs on the target floor plan.

\noindent\textbf{Reference Scene Clustering.}
To determine cluster assignment, each reference object $o_i^R$ is represented by a joint spatial-semantic feature vector,
%
%
\begin{equation}
\mathbf{x}_i^R = \left[\, \lambda\,\frac{\mathbf{p}_i^{xy}}{\|\mathbf{P}\|_F}
\;\oplus\; (1-\lambda)\,\frac{\hat{\phi}_i}{\|\hat{\Phi}\|_F} \,\right],
\label{eq1: cluster_assignment}
\end{equation}
where $\hat{\phi}_i$ is the $\ell_2$-normalized OpenShape~\cite{OpenShape} feature of $o_i^R$, and $\mathbf{P}$ and $\hat{\Phi}$ stack $\mathbf{p}_i^{xy}$ and $\hat{\phi}_i$ over all $N_R$ objects.
$\lambda \in [0,1]$ balances the two contributions, and $\oplus$ denotes vector concatenation.
Each block is normalized by the Frobenius norm $\|\cdot\|_F$, 
removing the scale disparity between 2D coordinates and high-dimensional semantic features.
Functional clusters $\mathcal{C}^R = \{\mathcal{C}_k^R\}_{k=1}^K$ are formed by applying agglomerative clustering over $\{\mathbf{x}_i^R\}$~\cite{ATT}. 
For each decomposed cluster $\mathcal{C}_k^R$, we compute its spatial centroid $\boldsymbol{\mu}_k^R \in \mathbb{R}^2$, representing member objects by their poses relative to $\boldsymbol{\mu}_k^R$. 
Additionally, we derive its axis-aligned spatial extent $\mathbf{s}_k^R \in \mathbb{R}^2$ and define its anchor orientation $\theta_k^R$ from the principal objects such as sofa, bed or desk. 

\noindent\textbf{Cluster Placement.}
We employ a Transformer~\cite{Transformer}-based placement network to predict target cluster poses $\mathbf{P}_{\mathcal{C}}^T = \{\mathbf{p}_{\mathcal{C}_k}^T\}_{k=1}^K$ on the floor plan $\mathcal{F}_T$. 
The target floor plan boundary is sampled as a set of 2D boundary points paired with inward unit normals, $\{(\mathbf{q}_l, \mathbf{n}_l)\}_{l=1}^{L}$, which a shallow Transformer encoder turns into a sequence of architectural context tokens $\mathbf{F}_{\mathcal{F}}$, one per boundary point, so that a cluster can attend to an individual wall rather than to a single pooled descriptor. 
Each cluster query token $\mathbf{v}_k$ concatenates an aggregation of the categories of its member objects, the cluster's current pose, and its spatial extent, 

\begin{equation}
\begin{aligned}
\mathbf{v}_k &= \big[\, \mathbf{u}_k \oplus \gamma(\mathbf{s}_k^R) \oplus \gamma(\tilde{\mathbf{p}}_{\mathcal{C}_k}^T) \,\big], \\
\mathbf{u}_k &= \frac{1}{|\mathcal{C}_k^R|}\sum_{i \in \mathcal{C}_k^R} \psi(\mathbf{c}_i^R),
\end{aligned}
\label{eq2: cluster_token}
\end{equation}
%
where $\mathbf{c}_i^R$ denotes the one-hot category vector of object $i$ embedded via per-object MLP $\psi$, 
$\tilde{\mathbf{p}}_{\mathcal{C}_k}^T$ is the current pose estimate during iterative denoising, and  $\gamma(\cdot)$ denotes a fixed positional encoding.
The placement network processes cluster tokens via self-attention across functional clusters and cross-attention over the floor plan tokens $\mathbf{F}_{\mathcal{F}}$, iteratively denoising cluster layout predictions. 
Finally, predicted cluster layouts undergo post-processing to resolve inter-cluster collisions, clamp extents within boundary polygon $\mathcal{F}_T$, and reserve clearance zones near openings $\mathcal{B}_T$.
Architectural details of our network are provided in the supplementary material.

\subsection{Cross-Scene Object Placement}
\label{subsec:object_placement}
While cluster-level placement establishes group locations, determining object-level arrangements is not straightforward.
Since no instance-level correspondence exists between reference and target object sets, the model must infer which target item maps to which reference counterpart.
Even where such a match is evident, target objects cannot simply inherit reference poses, since they differ from their reference counterparts in category and count, and must be re-derived against the target's own geometry. 
We resolve these challenges by conditioning placement on dense semantic representation of foundational features extracted from the transferred exemplar.

\noindent\textbf{Scene Context Tokens.}
Point clouds of reference objects are transformed into the target space based on predicted cluster poses and merged with the target floor point cloud. 
This combined point set is encoded using a pretrained Concerto~\cite{Concerto} point cloud feature extractor, and downsampled via farthest point sampling to obtain a set of spatial scene context tokens $\mathbf{F}_{\mathcal{S}}$. 
These scene tokens capture the spatial configuration currently occupying the target room, ensuring that subsequent object placement is conditioned on target-specific spatial geometry. 
	
\noindent\textbf{Object Tokens and Cross-Attention.}
Each target object $o_j^T$ is parameterized as a feature query token $\mathbf{h}_j^T$ by combining its semantic, geometric, and spatial properties,
\begin{equation}
\mathbf{h}_j^T = \left[\, 
\psi(\mathbf{c}_j^T) \oplus 
\gamma\!\left(\mathbf{s}_j^T\right) \oplus
\mathbf{W}_p \phi_{\mathrm{OS}}(o_j^T) \oplus \gamma\!\left(\tilde{\mathbf{p}}_j^T\right) \,
\right],
\label{eq3: object_token}
\end{equation}
where $\mathbf{c}_j^T$ is the one-hot category vector, $\mathbf{s}_j^T \in \mathbb{R}^2$ denotes the floor plane extent of the bounding box, and $\mathbf{W}_p$ is a trainable projection matrix mapping OpenShape~\cite{OpenShape} features $\phi_{\mathrm{OS}}(o_j^T)$ into the joint query space. 
The decoder denoises poses iteratively, so the token also carries the object's current pose $\tilde{\mathbf{p}}_j^T$, which lets each object be distinguished by where it currently stands rather than by category alone. 
A single additional token, obtained by encoding the target boundary points with a learnable lightweight point encoder~\cite{pointnet, Pointnet++}, supplies the room outline alongside the object tokens.
Within the layout decoder, target object tokens $\mathbf{h}_j^T$ cross-attend to the scene tokens $\mathbf{F}_{\mathcal{S}}$. 
Each scene token is the sum of its projected Concerto~\cite{Concerto} feature and a fixed positional encoding of the point it was sampled at.
This cross-attention mechanism enables each target object to identify spatial contexts in the transferred reference configuration that are analogous to its own 3D shape. 
Consequently, placement is conditioned not only on object category and bounding dimensions, but also on where semantically compatible structural configurations exist in the exemplar.
Using the attention-augmented features, the layout decoder predicts poses $\mathbf{P}_T = \{\mathbf{p}_j^T\}_{j=1}^{N_T}$ for target objects, which are updated through iterative denoising. 
\begin{figure}
  \centering
  \includegraphics[width=\linewidth]{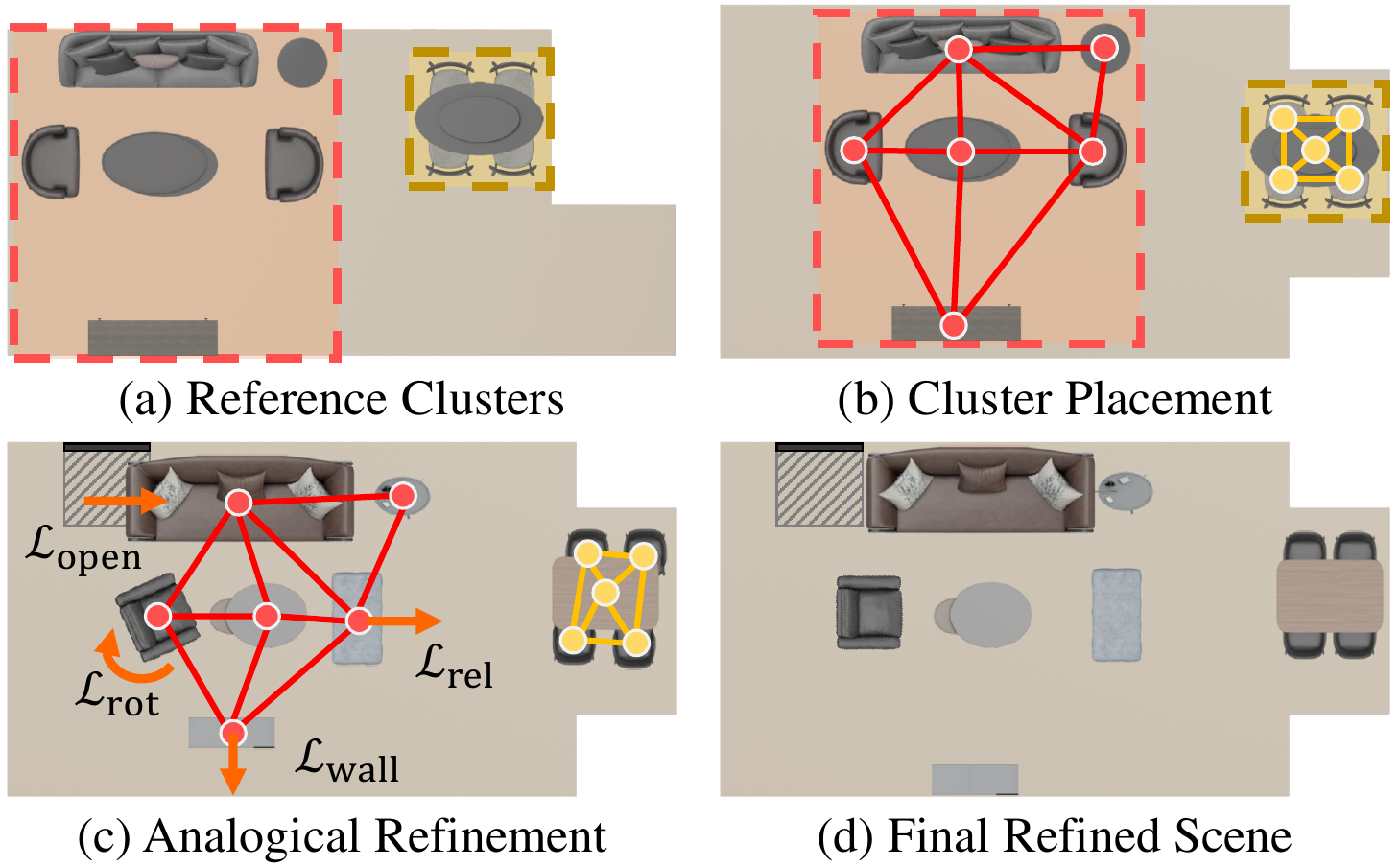}
  \caption{\textbf{Analogical Refinement.} Reference clusters (a) are placed on the target floor plan (b). After initial object arrangement, we apply graph matching followed by analogical refinement (c), yielding the final arrangement (d).
 }
  \label{fig:analogical_refinement}
\end{figure}
%

\subsection{Analogical Refinement}
\label{subsec:analogical_refinement}
While the Transformer~\cite{Transformer} layout decoder captures inter-object relations implicitly via self-attention, it is supervised only on per-object poses, so pairwise relations are never constrained directly and their errors compound across the scene.
The analogical refinement stage (\cref{fig:analogical_refinement}) corrects these relational discrepancies by directly extracting pairwise spatial constraints from the reference exemplar and re-establishing its wall and opening clearances against the target room, all under strict physical validity. 
This stage has no learnable parameters, with its constraints derived directly from the exemplar.

\noindent\textbf{Object Correspondence.}
Since reference and target inventories differ in count and category composition, instance-wise correspondence cannot be assumed, and must be explicitly established. 
Target objects first inherit cluster labels from the placed reference clusters by nearest neighbor propagation, which makes the cluster-level correspondence an identity map by construction and confines matching to within each cluster pair.
Within each matched cluster pair, object correspondences are established via reweighted random walk graph matching~\cite{GraphMatching, ATT}. Graph nodes encode OpenShape~\cite{OpenShape} features $\phi_{\mathrm{OS}}$, while graph edges capture pairwise spatial distances in cluster relative coordinates. 
Applying the Hungarian algorithm converts soft matching probabilities into a hard binary correspondence matrix $\mathbf{M} \in \{0, 1\}^{N_T \times N_R}$ along with matching confidence scores.

\noindent\textbf{Analogical Pose Refinement.}
Given the established correspondence matrix $\mathbf{M}$, target object poses $\mathbf{P}_T = \{\mathbf{p}_j^T\}_{j=1}^{N_T}$ are optimized to simultaneously preserve relative spatial structures from the exemplar and adapt to the architectural bounds of the target room.
The overall objective function $\mathcal{L}_{\mathrm{refine}}$ balances these relational and architectural constraints of the target floor plan,
\begin{equation}
\mathcal{L}_{\mathrm{refine}} = \mathcal{L}_{\mathrm{rel}} + \lambda_{\mathrm{rot}} \mathcal{L}_{\mathrm{rot}} + \lambda_{\mathrm{wall}} \mathcal{L}_{\mathrm{wall}} + \lambda_{\mathrm{open}} \mathcal{L}_{\mathrm{open}},
\label{eq4: refine_loss}
\end{equation}
where scalar hyperparameter weights balance the contribution of each term.
The relational term $\mathcal{L}_{\mathrm{rel}}$ pins each matched target object to the offset its counterpart held from the anchor of its own functional cluster in the reference scene,
%
\begin{equation}
\mathcal{L}_{\mathrm{rel}} = \frac{1}{|\mathcal{V}|}\sum_{j \in \mathcal{V}}
w_j \big\| \mathbf{d}_j - \mathbf{R}\,\mathbf{d}_j^R \big\|_2^2,
\label{eq5: relative_loss}
\end{equation}
where $\mathbf{d}_j = \mathbf{p}_j^{xy} - \mathbf{p}_{a(j)}^{xy}$ is the offset of target object $j$ from the anchor object $a(j)$ of its cluster, and
$\mathbf{d}_j^R = \mathbf{p}_{\pi(j)}^{R,xy} - \mathbf{p}_{\pi(a(j))}^{R,xy}$ is the corresponding offset in the reference, with $\pi(j)$ the matched reference index. 
$\mathbf{R}$ is the 2D cluster placement rotation, $w_j$ is the matching confidence, and $\mathcal{V}$ is the set of matched objects.
In addition to pairwise relative displacements, faithful transfer also requires preserving orientations and wall alignments, while the target room imposes its own boundary and openings.
First, $\mathcal{L}_{\mathrm{rot}}$ penalizes directional deviations from reference orientations $\theta_{\pi(j)}^R$.
Second, to preserve wall-adjacent layouts under changing room geometry, $\mathcal{L}_{\mathrm{wall}}$ enforces the reference signed clearance between object faces and walls in $\mathcal{S}_R$ onto the target boundary $\mathcal{F}_T$.
Finally, to guarantee spatial accessibility in the new environment, $\mathcal{L}_{\mathrm{open}}$ accumulates the squared penetration of object bounding boxes into walk-through clearance zones near target openings $\mathcal{B}_T$.
See the supplementary material for full formulation of each term and visualization of their individual effects.

Target poses $\mathbf{p}_j^T = (x_j^T, y_j^T, \theta_j^T)$ are optimized via gradient descent where physical validity is then enforced by projection rather than by penalty. 
Object poses are clamped inside the floor plan polygon before and after optimization, with object point cloud overlaps resolved by translating overlapping pairs along a separating axis.
Treating these constraints as projections keeps them exact, so a converged layout is collision-free and within bounds by construction. 
Objects with poor matching confidence scores or unresolvable physical collisions trigger automated pruning.

\section{Experiments}
\label{sec:experiments}
\begin{figure*}[t]
    \centering
    \includegraphics[width=\linewidth]{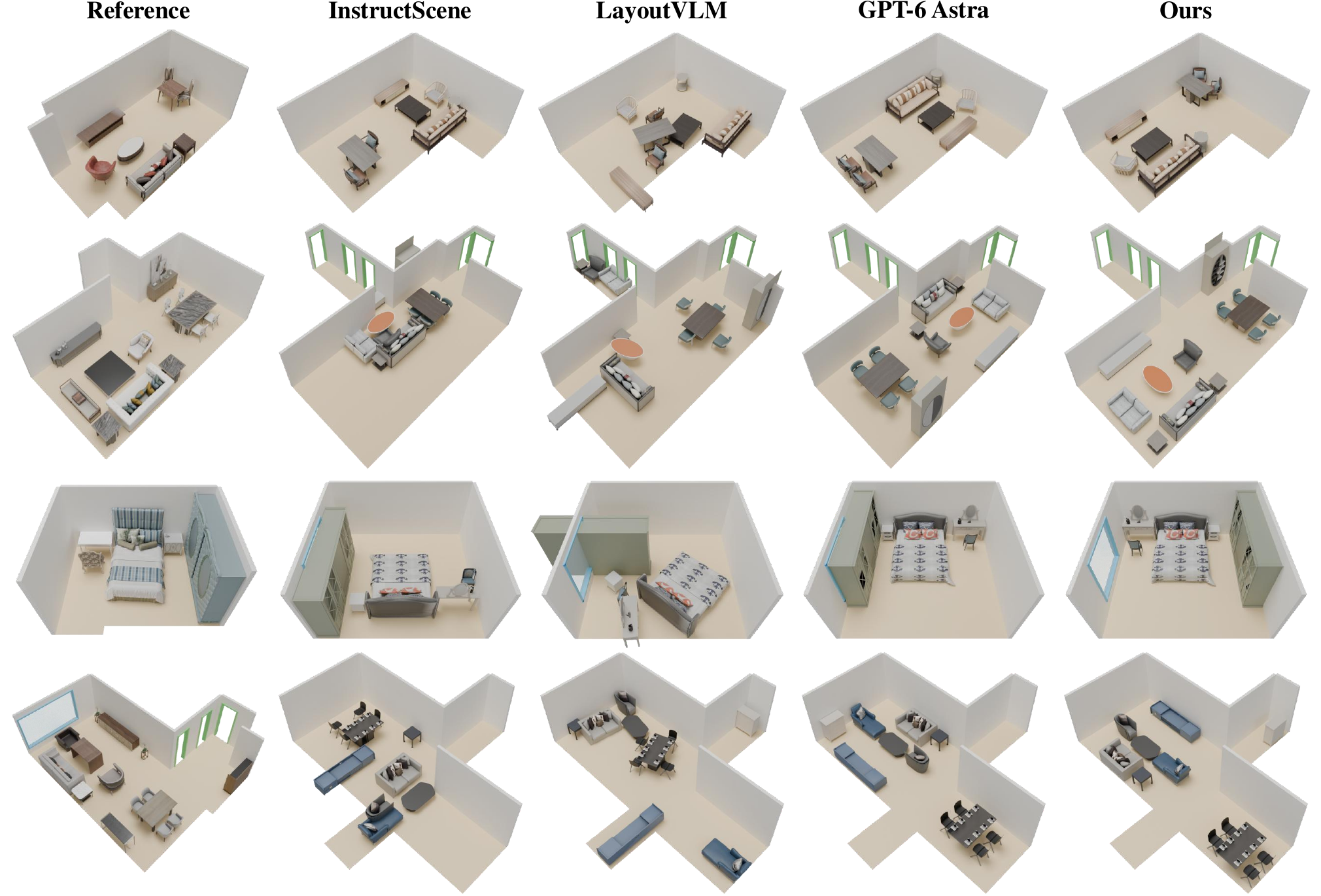}
    \vspace{-0.75em}
    \caption{\textbf{Qualitative Comparison.} Our hierarchical framework preserves fine-grained spatial relationships across diverse room geometries while strictly respecting physical constraints.}
    \label{fig:main_results}
    \vspace{-0.75em}
\end{figure*}
\subsection{Implementation Details}
\label{subsec:implementation_details}
\noindent\textbf{Training.}
Because ground truth arrangements do not exist for arbitrary reference and target scene pairs, direct supervision on the retargeting task is infeasible.
We therefore train both networks on a layout recovery task, perturbing ground-truth poses and supervising the networks to restore them.
Both cluster placement and layout decoders are trained using a mean squared error loss paired with a light $L_1$ penalty on concatenated position and orientation vectors.
Both are trained with Adam at learning rate $1\times10^{-4}$, and batch size 128 for 50k iterations (about two days).
The optimizer settings follow LEGO-Net~\cite{Lego-net} without modification to avoid bias from hyperparameter tuning.

\noindent\textbf{Inference.}
During inference, both networks employ an iterative denoising process. The cluster placement network denoises the cluster poses, starting from uniform random noise sampled within the normalized room coordinate range of $[-1, 1]^2$. 
Subsequently, the layout decoder denoises the target inventory poses, initialized from object poses randomly sampled from the same coordinate space. 
These iterative updates utilize decayed step sizes and annealed Langevin dynamics, with complete sampling schedules detailed in the supplementary material.

\noindent\textbf{Matching and Analogical Refinement.}
For clustering, we employ Ward linkage with $\lambda = 0.5$ in \cref{eq1: cluster_assignment}, targeting seven objects per cluster.
Analogical refinement runs for at most 100 iterations with
$(\lambda_{\mathrm{rot}}, \lambda_{\mathrm{wall}}, \lambda_{\mathrm{open}}) = (1.0, 0.5, 0.5)$.
Retargeting a scene takes 6.9\,s end-to-end, on a single NVIDIA GeForce RTX 4090 GPU paired with an AMD Ryzen 7 7700 CPU, which serves as the execution environment for all timing evaluations.

\subsection{Datasets \& Baselines}
\label{subsec:datasets}
\noindent\textbf{Datasets.} We train and evaluate on living room and bedroom scenes from 3D-FRONT~\cite{3D-Front} with assets from 3D-FUTURE~\cite{3D-Future}.
Training operates on individual rooms rather than paired scenes, using each room's unperturbed layout
as a recovery target: 9{,}352 living rooms and 22{,}672 bedrooms after axis-aligned rotation augmentation, with 2{,}348 and 896 held out.
For retargeting evaluation, we construct 50 living room and 25 bedroom pairs from the held-out split, each pairing a reference room with a same-type target room that supplies its floor plan, room openings, and object inventory.
Pairs are formed automatically by inventory similarity, since transferring structure between disparate inventories lacks meaningful correspondence.

\noindent\textbf{Baselines.}
We compare against representative layout generation models: denoising models (LEGO-Net~\cite{Lego-net}, DiffuScene~\cite{Diffuscene}, MiDiffusion~\cite{Midiffusion}), a scene graph-guided model (InstructScene~\cite{Instructscene}), language-driven synthesis (Holodeck~\cite{Holodeck}, I-Design~\cite{I-design}, LayoutGPT~\cite{Layoutgpt}, LayoutVLM~\cite{LayoutVLM}), and native 3D generation (LaviGen~\cite{LaviGen}).
Except for LEGO-Net and MiDiffusion, which see only the target floor plan,
these methods receive the reference scene as text.
To narrow the information gap between text and our 3D input, we further evaluate variants given the exact reference layout (category, size, position, and orientation of every object): LayoutVLM$^\dagger$, whose VLM receives the layout alongside the text, and InstructScene$^\dagger$, whose relation graph is extracted from the layout instead of generated from text.
We also evaluate a frontier general-purpose model, GPT-6 Astra~\cite{Astra}, prompted with the same layout and instruction.
Since our pipeline prunes 1.6 objects per scene on average, all baselines are
evaluated on the same pruned inventory, and Tab.~\ref{tab:ablation}~(d) isolates the effect of pruning.

%
\ifdefined\hdw\else\newlength{\hdw}\fi
\ifdefined\hdtmp\else\newlength{\hdtmp}\fi
\providecommand{\hdmax}[1]{\settowidth{\hdtmp}{#1}\ifdim\hdtmp>\hdw\setlength{\hdw}{\hdtmp}\fi}
\providecommand{\hd}[1]{\makebox[\hdw][c]{#1}}
\begin{table}[t]
\centering
\footnotesize
\setlength{\tabcolsep}{2pt}
\setlength{\hdw}{0pt}\hdmax{Overall}\hdmax{PSA $\uparrow$}\hdmax{Pos. $\uparrow$}
\resizebox{\linewidth}{!}{
\begin{tabular}{@{}l|c|cccccc@{}}
\toprule
\multirow{2}{*}[-2pt]{Method} & \multirow{2}{*}[-2pt]{Ref.} & \multicolumn{2}{c}{Physical} & \multicolumn{2}{c}{Semantic} & \multicolumn{1}{c}{Overall} & \multicolumn{1}{c}{Fidelity} \\
\cmidrule(lr){3-4} \cmidrule(lr){5-6} \cmidrule(lr){7-7} \cmidrule(lr){8-8}
 & & \hd{CF $\uparrow$} & \hd{IB $\uparrow$} & \hd{Pos. $\uparrow$} & \hd{Rot. $\uparrow$} & \hd{PSA $\uparrow$} & iRecall $\uparrow$ \\
\midrule
LEGO-Net~\cite{Lego-net} & - & 84.3 & 68.4 & 40.3 & 42.6 & 24.3 & 42.8 \\
DiffuScene~\cite{Diffuscene} & Text & 83.1 & 50.0 & 39.3 & 39.0 & 17.7 & 38.4 \\
MiDiffusion~\cite{Midiffusion} & - & 85.8 & 62.8 & 44.5 & 44.9 & 25.3 & 41.0 \\
InstructScene~\cite{Instructscene} & Text & 87.7 & 51.7 & \underline{56.7} & 54.8 & 24.3 & 46.8 \\
Holodeck~\cite{Holodeck} & Text & 77.2 & 64.1 & 43.8 & 43.9 & 21.0 & 41.8 \\
I-Design~\cite{I-design} & Text & 75.7 & 52.1 & 38.5 & 39.7 & 13.8 & 33.7 \\
LayoutGPT~\cite{Layoutgpt} & Text & 81.3 & 45.4 & 47.7 & 45.4 & 18.5 & 37.9 \\
LayoutVLM~\cite{LayoutVLM} & Text & 87.2 & 86.8 & 49.9 & 46.7 & 37.5 & 42.5 \\
LaviGen~\cite{LaviGen} & Text & \underline{95.9} & 96.4 & 43.3 & 44.4 & 46.9 & 44.4 \\
\midrule
InstructScene$^\dagger$~\cite{Instructscene} & Layout & 94.1 & 57.1 & 52.4 & 53.2 & 32.6 & 49.0 \\
LayoutVLM$^\dagger$~\cite{LayoutVLM} & Layout & 89.3 & 87.7 & 49.3 & 48.7 & 42.8 & 41.2 \\
GPT-6 Astra~\cite{Astra} & Layout & 94.5 & \underline{98.3} & 55.6 & \underline{54.9} & \underline{58.7} & \underline{56.5} \\
\midrule
Ours & 3D scene & \textbf{97.9} & \textbf{99.8} & \textbf{62.8} & \textbf{61.9} & \textbf{63.5} & \textbf{64.3} \\
\bottomrule
\end{tabular}
}
\vspace{-0.5em}
\caption{\textbf{Quantitative Comparison.} Our method outperforms recent 3D layout generation baselines across all evaluation metrics. 
The Ref. column gives how each method receives the reference scene, a dash marks methods that receive only the target floor plan, and $^\dagger$ marks variants given the exact reference layout. 
Best and second-best results are formatted in \textbf{bold} and \underline{underlined}.}
\label{tab:comparison}
\vspace{-0.5em}
\end{table}
\subsection{Comparative Studies}
\label{subsec:comparative_studies}
Since no ground-truth layouts exist for target rooms, we evaluate performance across physical, semantic, and overall metrics following LayoutVLM~\cite{LayoutVLM}.
For each reference room, we build (subject, relation, object) triplets covering directional adjacency, facing, alignment, and wall contact.
These triplets are verbalized into a cached instruction that is shared by every method as baseline input and as the judge's layout criteria (see the supplementary material for prompt details).
\textbf{Physical Plausibility} is measured by Collision Free (CF) and In Boundary (IB) rates.
\textbf{Semantic Coherency} evaluates Positional (Pos) and Rotational (Rot) coherence via GPT-5.6~\cite{GPT-5} ratings on rendered scenes.
Finally, \textbf{Physically Grounded Semantic Alignment (PSA)} serves as the overall score, combining GPT-5.6 semantic evaluations with physical validity.
To evaluate relationship preservation, we adopt \textbf{Instruction Recall (iRecall)} from InstructScene~\cite{Instructscene}, which measures the fraction of reference spatial triplets realized in the generated geometry.

\Cref{tab:comparison} and \cref{fig:main_results} summarize the quantitative and qualitative evaluations, respectively.
Our framework achieves the best performance across every metric, and the results reveal a clear divide among existing baselines.
Methods that produce physically valid layouts fail to reproduce the exemplar: LaviGen~\cite{LaviGen} attains the highest physical scores among text-conditioned baselines yet yields only 44.4 iRecall.
Conversely, methods prioritizing semantic structure severely compromise physical validity.
InstructScene~\cite{Instructscene} achieves the top coherency and iRecall among text-conditioned baselines, but stays within room boundaries in barely half of its generations (51.7 IB).
LayoutVLM~\cite{LayoutVLM} serves as an informative intermediate case, where its differentiable optimization stage restores physical validity, yet its iRecall of 42.5 still falls below that of InstructScene.
No text-conditioned baseline simultaneously satisfies physical plausibility and structural fidelity, reflecting the information bottleneck of text and symbolic abstractions.

Supplying the exact reference layout narrows but does not close this gap.
It lifts InstructScene$^\dagger$ to 49.0 iRecall and LayoutVLM$^\dagger$ to 42.8 PSA, while GPT-6 Astra~\cite{Astra} becomes the strongest baseline (58.7 PSA, 56.5 iRecall).
Rather than transferring the exemplar, however, Astra composes a plausible generic layout around the reference's coarse grouping, and finer details such as relative offsets, orientations, and wall clearances are lost as shown in~\cref{fig:main_results}.
By operating directly on 3D features, our method resolves this tension, advancing iRecall from 56.5 to 64.3 and PSA from 58.7 to 63.5 over the strongest baseline.

\subsection{Ablation Studies}
\label{subsec:ablation_studies}
\ifdefined\hdw\else\newlength{\hdw}\fi
\ifdefined\hdtmp\else\newlength{\hdtmp}\fi
\providecommand{\hdmax}[1]{\settowidth{\hdtmp}{#1}\ifdim\hdtmp>\hdw\setlength{\hdw}{\hdtmp}\fi}
\providecommand{\hd}[1]{\makebox[\hdw][c]{#1}}
\begin{table}[t]
\centering
\footnotesize
\setlength{\tabcolsep}{2pt}
\setlength{\hdw}{0pt}\hdmax{Overall}\hdmax{PSA $\uparrow$}\hdmax{Pos. $\uparrow$}
\resizebox{\linewidth}{!}{
\begin{tabular}{@{}l|cccccc@{}}
\toprule
\multirow{2}{*}[-2pt]{Method} & \multicolumn{2}{c}{Physical} & \multicolumn{2}{c}{Semantic} & \multicolumn{1}{c}{Overall} & \multicolumn{1}{c}{Fidelity} \\
\cmidrule(lr){2-3} \cmidrule(lr){4-5} \cmidrule(lr){6-6} \cmidrule(lr){7-7}
 & \hd{CF $\uparrow$} & \hd{IB $\uparrow$} & \hd{Pos. $\uparrow$} & \hd{Rot. $\uparrow$} & \hd{PSA $\uparrow$} & iRecall $\uparrow$ \\
\midrule
(a) w/o Cluster Placement & 94.3 & 94.6 & 49.4 & 47.8 & 47.7 & 51.6 \\
(b) w/o Analogical Refinement & 93.8 & 99.2 & 58.5 & 56.0 & 54.8 & 56.8 \\
(c) w/o Physical Constraints & 88.4 & 92.7 & 57.8 & 56.8 & 54.7 & 61.5 \\
(d) w/o Pruning & \underline{96.8} & 98.7 & \underline{61.2} & \underline{60.6} & \underline{62.9} & 60.1 \\
(e) w/o Object Arrangement & 94.4 & 98.9 & 54.7 & 54.6 & 52.1 & 56.4 \\
(f) Refinement-only & 90.1 & 91.9 & 40.9 & 36.2 & 36.1 & 40.5 \\
\midrule
(g) w/o OpenShape & 96.2 & \underline{99.6} & 56.0 & 55.9 & 53.8 & 59.5 \\
(h) w/o Concerto & 95.6 & \textbf{99.8} & 52.2 & 51.7 & 51.1 & 54.2 \\
(i) w/o Floorplan & 95.9 & 97.0 & 57.3 & 55.9 & 56.2 & \underline{61.8} \\
\midrule
Ours (full) & \textbf{97.9} & \textbf{99.8} & \textbf{62.8} & \textbf{61.9} & \textbf{63.5} & \textbf{64.3} \\
\bottomrule
\end{tabular}
}
\vspace{-0.5em}
\caption{\textbf{Ablation Study.} Pipeline components (a--f) and conditioning signals (g--i) are evaluated. Best and second-best results are formatted in \textbf{bold} and \underline{underlined}.}
\label{tab:ablation}
\vspace{-0.5em}
\end{table}
\begin{figure*}[t]
    \centering
    \includegraphics[width=\linewidth]{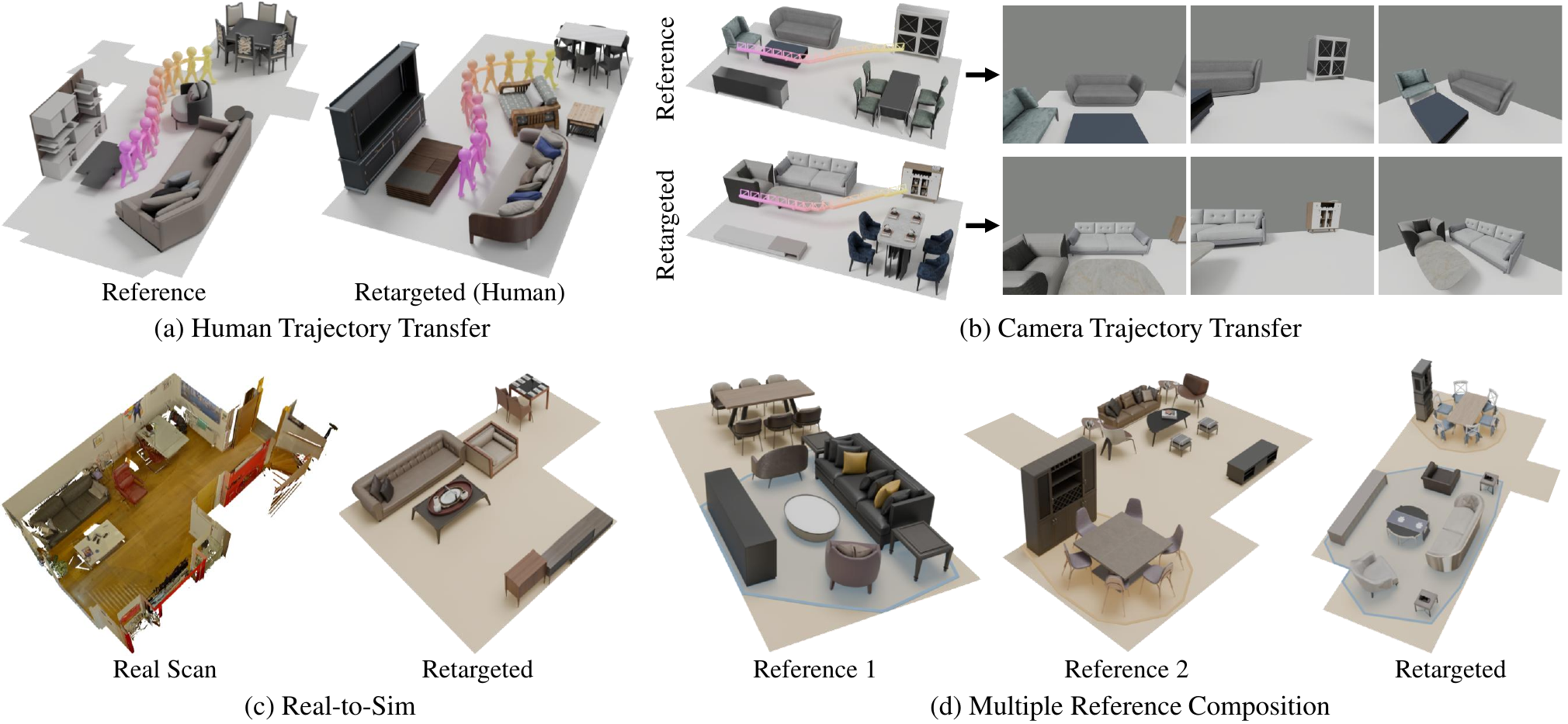}
    \vspace{-1.5em}
    \caption{\textbf{Downstream Applications.} Our framework enables various applications such as (a) human and (b) camera trajectory transfer, (c) real-to-sim and (d) multiple reference composition.}
    \label{fig:application}
    \vspace{-0.75em}
\end{figure*}
\Cref{tab:ablation} validates the necessity of each component across pipeline stages and conditioning signals.
In variant (a), Concerto features are extracted from the reference scene in its original coordinate frame and fed directly to the layout decoder, so the target inventory is arranged against the raw exemplar.
This represents the most direct alternative to learned cluster placement, and the 12.7 drop in iRecall confirms that structural transfer must occur before object-level arrangement.
Analogical refinement (b) provides fine-grained relational alignment, increasing iRecall by 7.5.
Removing physical constraints (c) degrades CF by 9.5 and IB by 7.1, confirming their role in enforcing geometric boundaries while preserving relational fidelity.
Disabling pruning (d) keeps PSA nearly unchanged (62.9) but lowers iRecall by 4.2, as unmatched objects remain as obstacles, breaking some reference relations.
Replacing the layout decoder with Hungarian matching (e), which assigns each target object the pose of the transplanted reference object with the same category and most similar OpenShape feature, lowers iRecall to 56.4.
Refinement alone (f) reaches only 40.5, showing that refinement requires a reasonable initialization rather than replacing it.

Regarding conditioning signals, global scene context dominates isolated asset representations.
Excluding Concerto scene features (h) incurs a steep 10.1 drop in iRecall, compared to a 4.8 decrease when removing OpenShape object features (g), demonstrating that scene-level spatial context is more critical for analogical placement than asset geometry.
Finally, omitting floor plan boundaries (i) lowers IB by 2.8 and PSA by 7.3, while relational fidelity stays largely intact at 61.8 iRecall.
In summary, while cluster placement and 3D foundation features dominate overall structural fidelity, analogical refinement ensures precise relational alignment and geometric constraints strengthen physical plausibility.
See the supplementary material for visual comparisons.

\subsection{Applications}
\label{subsec:applications}
Our framework naturally supports various downstream scenarios without fine-tuning.
We demonstrate three key applications below and detail same-inventory relocation in supplementary material.

\noindent\textbf{Analogical Trajectory Transfer.}
As illustrated in \cref{fig:application}(a, b), because spatial relations are faithfully preserved in the target scene, trajectories authored in a reference scene carry over to target environments without manual re-authoring.
Waypoints attached to reference objects map directly to their target counterparts~\cite{ATT}.
This enables the direct transfer of human interaction paths~\cite{lim_human_interaction} as well as 3D camera trajectories like those generated by HouseTour~\cite{Housetour}, preserving navigational context across varied geometries.

\noindent\textbf{Real-to-Sim.}
While standard reconstruction pipelines convert physical scans into single simulator-ready replicas~\cite{SimRecon, Litereality, MetaScenes}, Scene Retargeting instantiates a real ScanNet++~\cite{ScanNet++} scan across different synthetic floor plans with object inventories as illustrated in~\cref{fig:application}(c).
Consequently, a single real-world capture yields a diverse family of simulation environments rather than a single rigid replica, enabling automated environment generation from physical scans without requiring manual 3D modeling.

\noindent\textbf{Multiple Reference Composition.}
Since transfer operates at the functional cluster level, functional groups drawn from multiple independent reference rooms seamlessly assemble into a single target layout as depicted in~\cref{fig:application}(d).
Unlike existing layout generation methods~\cite{Diffuscene, Midiffusion, Instructscene} whose global conditioning cannot integrate configurations from distinct source rooms, our localized 3D cluster formulation enables multi-source composition.

\section{Conclusion} 
\label{sec:conclusion}
In this paper, we formalize the task of \textbf{Scene Retargeting} and propose a three-stage framework to transfer the functional arrangement of a reference 3D scene onto empty target rooms with distinct floor plans and mismatched object inventories.
Our approach first positions functional clusters using a learned placement network, and subsequently leverages 3D features from pretrained foundation models to guide object-level arrangements within those clusters.
An analogical refinement stage further enforces fine-grained relational fidelity, physical plausibility, and navigational affordances under strict geometric constraints.
Evaluations demonstrate that our method outperforms existing baselines across physical validity and semantic coherence metrics, while enabling versatile downstream applications.

Our framework focuses on 2D floor plane poses and assumes comparable inventories, leaving vertical stacking and severe object category mismatches outside its scope. 
Extending the formulation to 3D vertical structures, and combining exemplar
transfer with generative or retrieval-based completion where a target room cannot realize the reference's structure, remain promising future directions.

\section*{LLM Usage Statement}
We used Claude (Opus 5.5) to edit the text of this manuscript, specifically for grammar correction and sentence-level rewriting for clarity and concision.
Claude also assisted in adapting released baseline code to our target object inventories and debugging code crashes. 
For the experiments and evaluation, GPT-5.6 was utilized to rewrite templated relation clauses into instructions for text-conditioned baselines and to serve as an automated judge for semantic coherency and PSA scores.
All text and code produced or refined with AI assistance were thoroughly reviewed and verified line-by-line by the authors. 
The authors retain full responsibility for all content, claims, and implementations in this work.

{
    \small
    \bibliographystyle{ieeenat_fullname}
    \bibliography{main}
}

\clearpage
\appendix
\maketitlesupplementary

\section{Network Architecture}
\begin{figure*}[t]
    \centering
    \includegraphics[width=\linewidth]{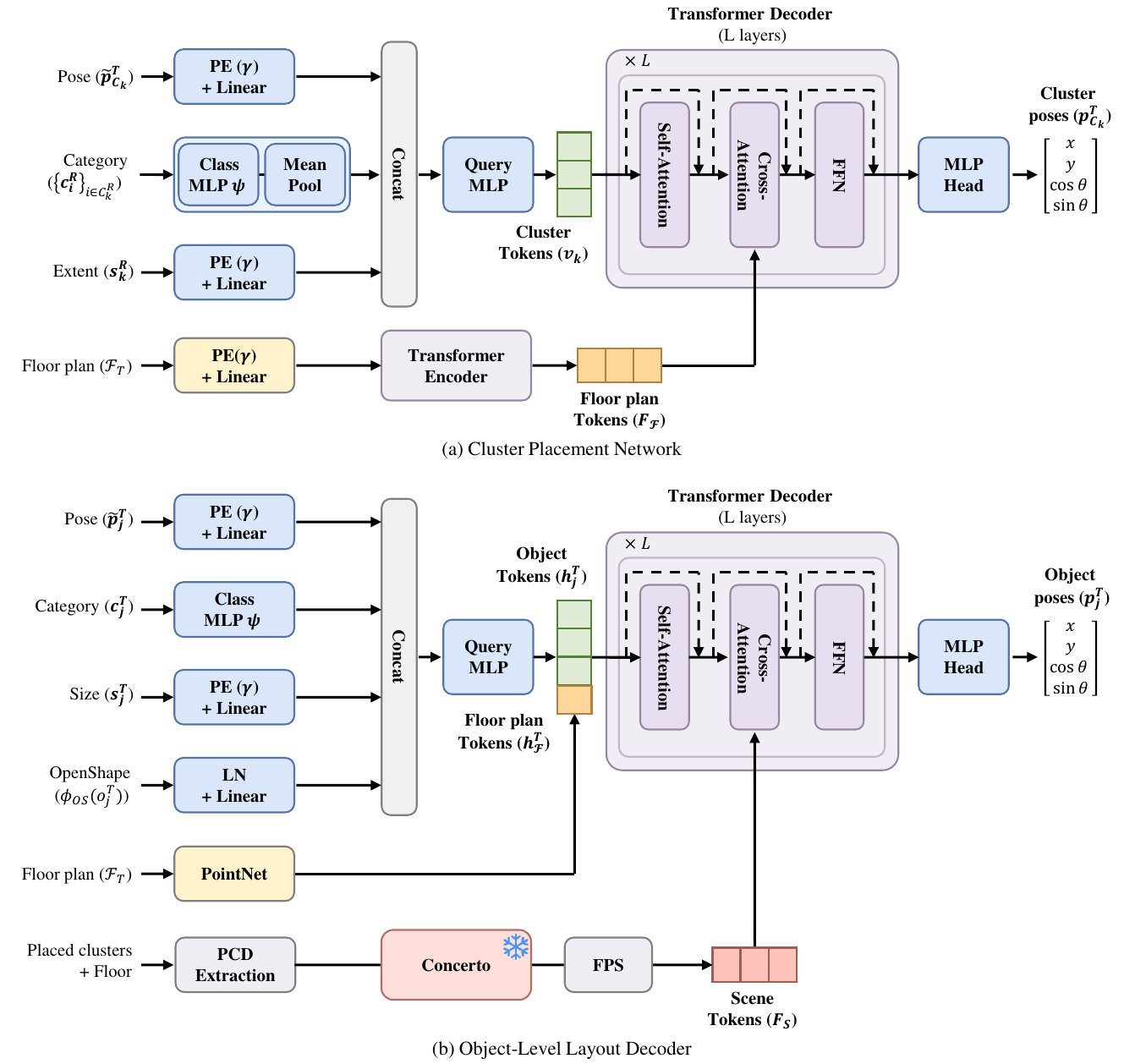}
    \caption{\textbf{Network Architecture.} (a) Cluster placement network for predicting cluster poses conditioned on the target floor plan. (b) Object-level layout decoder for predicting object poses from the placed reference objects and target floor plan.}
    \label{fig:network_architecture}
\end{figure*}

Both learned stages leverage Transformers~\cite{Transformer}, where each receives a perturbed layout alongside its conditioning and predicts the clean one. 
All coordinates, extents and angles are embedded with fixed sinusoidal encodings $\gamma(\cdot)$ rather than learned ones. 
The OpenShape~\cite{OpenShape} encoder $\phi_{\mathrm{OS}}$ and the Concerto~\cite{Concerto} encoder are frozen and used purely as feature extractors, so only the two networks above are trained. \cref{fig:network_architecture} shows the two networks.

\noindent\textbf{Cluster Placement Network.}
Each cluster $\mathcal{C}^R_k$ is reduced to a single query token $\mathbf{v}_k$.
The one-hot category vectors $\mathbf{c}^R_i$ of its member objects $i \in \mathcal{C}^R_k$ pass through a two-layer MLP $\psi$ and are mean-pooled over the valid object slots into $\mathbf{u}_k$.
This is concatenated with encodings of the cluster's spatial extent $\gamma(\mathbf{s}^R_k)$, of its current pose estimate $\gamma(\tilde{\mathbf{p}}^T_{\mathcal{C}_k})$, and the concatenation is projected back to width $256$.
The target room enters as $L = 250$ boundary points $\{(\mathbf{q}_l, \mathbf{n}_l)\}_{l=1}^{L}$ carrying position and inward normal, encoded by a two-layer pre-norm Transformer~\cite{Transformer} encoder of width $256$ with four heads into $250$ spatial tokens $\mathbf{F}_{\mathcal{F}}$.
Four decoder layers of width $256$ with four heads and feed-forward width $1024$ then alternate self-attention across cluster queries $\{\mathbf{v}_k\}_{k=1}^{K}$ with cross-attention to those boundary tokens $\mathbf{F}_{\mathcal{F}}$, so a cluster is placed with reference to the room's geometry and to the other clusters at once.
An MLP head emits each cluster center's position and its orientation as $\mathbf{p}^T_{\mathcal{C}_k} \in \mathbf{P}^T_{\mathcal{C}}$.

\noindent\textbf{Object-level Layout Decoder.}
Each object token $\mathbf{h}^T_j$ concatenates a $16$-dimensional encoding of its size $\gamma(\mathbf{s}^T_j)$, a $64$-dimensional encoding of its category $\mathbf{c}^T_j$, and its $1280$-dimensional OpenShape~\cite{OpenShape} feature $\phi_{\mathrm{OS}}(o^T_j)$ projected to $64$ by $\mathbf{W}_p$, together with an encoding of its current pose estimate $\gamma(\tilde{\mathbf{p}}^T_j)$, and is projected to width $512$.
The point clouds of the reference objects $\{o^R_i\}_{i=1}^{N_R}$, placed at their cluster-assigned poses $\mathbf{P}^T_{\mathcal{C}}$, are concatenated with the point cloud of the target room's floor, and Concerto~\cite{Concerto} encodes this combined point cloud into a per-point feature field.
Farthest point sampling then selects $768$ of those points, which carry their features with them, and the retained $1536$-dimensional features are projected to $512$ to form the scene tokens $\mathbf{F}_{\mathcal{S}}$.
A PointNet~\cite{pointnet, Pointnet++} encoder over the same $250$ boundary points supplies one further token $\mathbf{h}^T_{\mathcal{F}}$ so that the floor plan $\mathcal{F}_T$ is visible to the decoder directly rather than only through the scene point cloud.
Six decoder layers of width $512$ with eight heads let the object tokens $\{\mathbf{h}^T_j\}_{j=1}^{N_T}$ attend to this memory, and an MLP head emits each object's position and orientation as $\mathbf{p}^T_j \in \mathbf{P}_T$.
%
\section{Training and Inference Details}
Our framework optimizes parameters for the cluster placement network and the
object-level layout decoder, while clustering, graph matching, and
analogical refinement operate entirely parameter-free.

\noindent\textbf{Self-supervised Layout Recovery.}
Because paired ground-truth arrangements do not exist for arbitrary reference and
target room pairs, direct supervision on retargeting is infeasible. 
We instead train both networks on a self-supervised layout recovery task, where each network reconstructs a clean layout from a synthetically perturbed one.
Following LEGO-Net~\cite{Lego-net}, the perturbation scale is drawn once per scene and applied independently to each object. 
Positions are displaced by $\mathcal{N}(0, \sigma_p^2)$ with $\sigma_p \sim |\mathcal{N}(0, 0.15^2)|$ in normalized room coordinates, and orientations by $\mathcal{N}(0, \sigma_a^2)$ with $\sigma_a \sim |\mathcal{N}(0, (\pi/4)^2)|$. 
Training operates on individual rooms, with each room's unperturbed layout serving as the recovery target. 
During training, scene tokens are built from objects of the same room in their original poses.
At inference, they are built from placed reference clusters, with object tokens drawn from the target inventory.
Since object tokens use OpenShape~\cite{OpenShape} features and scene tokens use Concerto~\cite{Concerto} features, the decoder cannot match the two using identical features. 
Instead, it learns to associate object shape with scene context, which carries over to objects of similar shape. 
Differences in object count and size are unseen during training and are handled by the correspondence and refinement steps in stage 3.

\noindent\textbf{Objectives.}
Both networks are supervised on the concatenated position and orientation vector with a squared error together with a small $L_1$ term at weight $0.07$,
the $L_1$ component sharpening convergence once the squared error has flattened.
The cluster placement network predicts the cluster centroid and anchor orientation, and the layout decoder predicts the pose of each object.
Orientation is represented as the two-vector $(\cos\theta, \sin\theta)$ rather than as an angle, which keeps the loss insensitive to the wrap-around at $\pm\pi$. The networks regress this vector directly, and the prediction is renormalized to unit length at inference.

\noindent\textbf{Optimizer.}
Both networks are trained with Adam using a batch size of 128 for 50k iterations with a learning rate of $1\times10^{-4}$ following the reference implementation~\cite{Lego-net}. 
We adopt these optimizer settings without modification, so that the comparison in Tab. 1 of the main manuscript does not reflect differences in training-time tuning effort.

\noindent\textbf{Dataset Augmentation.}
A separate model is trained for each room type, on the splits reported in
Sec. 4.2 of the main manuscript. Training uses the axis-aligned rotation augmentation of the preprocessed dataset, which rotates a room by a multiple of ninety degrees together with its floor plan, boundary samples and object poses.
Because an opening is parameterized as an interval on a boundary edge rather than as a coordinate, it survives this augmentation unchanged. 
The boundary representation seen during training carries position and normal alone, so openings reach the arrangement through the geometric constraints rather than as a learned conditioning signal.

\noindent\textbf{Inference.}
At inference, both networks perform iterative denoising from random initial layouts. 
Cluster poses and, subsequently, target object poses are sampled uniformly at random in the normalized room coordinate range $[-1, 1]^2$, with orientations sampled uniformly from $[-\pi, \pi)$. 
Neither the reference cluster centroids nor the original object poses of the target room are used. 
Although training perturbations are smaller, the decayed step size lets the networks reduce large initial errors gradually, as in LEGO-Net~\cite{Lego-net}. 
At iteration $t$, the layout updates toward the network prediction using a step size of $0.1/(1+0.005t)$, augmented by annealed Langevin noise scaled by $0.01 \cdot 0.9^{\lfloor t/10 \rfloor}$. Both stages run for at most $100$ steps, terminating early when layout displacement falls below $0.01$ in position and $0.005$ in $(\cos\theta,\sin\theta)$ orientation over three consecutive steps.

\section{Analogical Refinement}
\subsection{Objective Functions}
The relational term $\mathcal{L}_{\mathrm{rel}}$ in Eq. 5 of the main manuscript refines where a matched object stands relative to its cluster anchor. 
The remaining terms in Eq. 4 of the main manuscript constrain what that displacement alone leaves free.
Throughout, quantities measured on $\mathcal{S}_R$ are computed once before optimization and held fixed, so each term pulls the target layout toward a constant that the exemplar supplies.

\noindent\textbf{Orientation.} Faithful transfer requires a matched object to preserve its counterpart's orientation. 
To account for cluster rotation $\mathbf{R}$, $\mathcal{L}_{\mathrm{rot}}$ transforms the reference heading into the target frame:
\begin{equation}
\mathcal{L}_{\mathrm{rot}} = \frac{1}{|\mathcal{V}|}\sum_{j \in \mathcal{V}}
w_j \big\| \mathbf{e}(\theta_j^T) - \mathbf{R}\,\mathbf{e}(\theta_{\pi(j)}^R) \big\|_2^2,
\label{eq6: rotation_loss}
\end{equation}
where $\mathbf{e}(\theta) = (\cos\theta, \sin\theta)^\top$ denotes the unit heading
vector. Formulating the loss via 2D unit vectors, rather than raw angles $\theta$,
eliminates wrap-around discontinuities at $\pm\pi$ while strictly penalizing
$180^\circ$ orientation flips.

\noindent\textbf{Wall Clearance.}
Rather than absolute coordinates, relative wall clearances serve as the primary target for transfer. For instance, a wardrobe against a wall in $\mathcal{S}_R$ must stand against a wall in $\mathcal{F}_T$ even though the walls are located elsewhere.
Wall correspondence is established per cluster.
Specifically, each wall bounding $\mathcal{C}^R_k$ is matched to the target wall whose inward normal best agrees with it under $\mathbf{R}$ and whose span overlaps the placed cluster's projection; an object is tied only to walls matched for its cluster.
Object $j$ is assigned such a wall when its reference counterpart stood within $0.2$ of that wall's reference edge in normalized room coordinates.
At most two walls are retained per object, selected as the two closest in the reference scene. When two walls are kept, their normals must differ by at least $60^\circ$, ensuring a piece in a corner respects both adjacent walls while a piece in open floor remains unconstrained.
Let $g(\mathbf{p}, \mathbf{s}; e)$ denote the gap from the near face of a box of extent $\mathbf{s}$ at pose $\mathbf{p}$ to the line of wall $e$, signed positive when the box lies clear of the wall and negative when it penetrates.
With $\mathcal{W}$ denoting the set of resulting assignments and $e^R$ the reference wall from which $e$ was matched,
\begin{equation}
\mathcal{L}_{\mathrm{wall}} = \frac{1}{|\mathcal{W}|}\sum_{(j,e) \in \mathcal{W}}
\big( g_{j,e}^T - g_{j,e}^R \big)^2,
\label{eq7: wall_loss}
\end{equation}
where $g_{j,e}^T = g(\mathbf{p}_j^T, \mathbf{s}_j^T; e)$ and
$g_{j,e}^R = g(\mathbf{p}_{\pi(j)}^R, \mathbf{s}_{\pi(j)}^R; e^R)$ are the signed
clearances of object $j$ and its reference counterpart to their respective walls.
The optimization target thus matches the clearance the counterpart held, and the
constraint set is determined by the exemplar rather than where an object happens
to land.

\noindent\textbf{Openings.}
Each opening induces a walk-through clearance zone $Z_b$, defined as a rectangle spanning the opening's boundary interval and extending $0.75\,\mathrm{m}$ inward. 
This depth is specified in meters rather than normalized units because it is determined by human body dimensions rather than room scale.
For windows elevated above the floor, the penalty is restricted to objects whose vertical extent is tall enough to obstruct the window aperture.
The penalty is defined as
\begin{equation}
\mathcal{L}_{\mathrm{open}} = \frac{1}{|\mathcal{B}_T|}
\sum_{b} \sum_{j=1}^{N_T}
\delta(\mathbf{p}^T_j, \mathbf{s}^T_j; Z_b)^2,
\label{eq8: open_loss}
\end{equation}
where $\delta$ denotes the separating-axis penetration depth of the box of extent $\mathbf{s}^T_j$ at pose $\mathbf{p}^T_j$ into $Z_b$. 
Using penetration depth rather than a corner test reliably penalizes edge cases, such as a wide sideboard spanning a narrow doorway with all four corners landing outside the zone.
Open archways, which 3D-FRONT~\cite{3D-Front} annotates as holes rather than doors, are excluded from $\mathcal{B}_T$ and thus contribute neither a clearance zone nor a denominator term. 
Because archways are far wider than standard doors, reserving a walk-through zone across them would unnecessarily restrict usable floor space without improving navigability.

\subsection{Physical Constraints}
All target poses are optimized jointly via gradient descent on $\mathcal{L}_{\mathrm{refine}}$ (Eq. 4 of the main manuscript). 
Physical validity is enforced through direct projection rather than penalty terms. 
Objects are clamped within $\mathcal{F}_T$ both before and after optimization, and residual overlaps are resolved by translating the overlapping pair along a separating axis of their oriented bounding boxes. 
Because oriented boxes over-approximate object geometry, candidate overlaps are verified against the two objects' point clouds prior to displacement, with translation distances measured directly from these point clouds rather than from the boxes. 
Consequently, pairs whose boxes intersect while their point clouds do not, such as a chair placed under a table, remain untouched. 
Finally, objects with low matching confidence $w_j$ or unresolvable collisions are pruned.

\section{Evaluation Details}
\subsection{Relation Triplets}
\begin{table*}[t]
\centering
\small
\begin{tabular}{llll}
\toprule
Rule & Labels & Distance & Angle \\
\midrule
Adjacency & \texttt{in\_front\_of}, \texttt{behind}, & gap $<0.22$ & $\pm45^{\circ}$ quadrants \\
          & \texttt{beside\_left}, \texttt{beside\_right} & & \\
Facing    & \texttt{facing}       & $\leq 0.6$ & $\pm 9^{\circ}$ cone \\
Alignment & \texttt{aligned\_with} & $\leq 0.6$ & headings within $20^{\circ}$ \\
Wall      & \texttt{against\_wall} & $<0.12$ to boundary & N/A \\
\bottomrule
\end{tabular}
\caption{\textbf{Relation Extraction Rules.} Distances are in normalized room coordinates; the adjacency gap is center distance minus the combined half-extents of both objects.}
\label{tab:relation_rules}
\end{table*}
Both reference-derived metrics, iRecall and PSA, are built on a single rule-based pass over the reference scene's ground-truth layout. Four rules emit seven relation labels, listed with their thresholds in \cref{tab:relation_rules}. 
Adjacent pairs are classified by direction rather than by mere proximity, since a coffee table in front of a sofa and a side table flanking it are different arrangements that a single \emph{next to} would collapse. 
The direction is read in the facing frame of one of the two objects, choosing whichever frame places the other object closest to one of its four axes, measured by maximizing $|\cos 2\phi|$ over the bearing $\phi$.
The facing cone is deliberately narrow, as a wider one lets an object squarely facing its true target also sweep in a distant object along the same bearing.
Every relation holds between two objects, or between one object and the nearest wall, and never as an absolute compass direction or a fraction of the room's dimensions. 
This ensures the same relation remains meaningful in a target room of a different shape, allowing the extracted facts to serve both as the specification handed to the baselines and as the quantity we score. 
These thresholds were determined empirically on the validation split. 

\subsection{Instruction Synthesis}
The triplets become the natural-language instruction in two steps.
A deterministic verbalizer first writes one plain English clause per triplet, so that no model chooses what the specification contains.
A language model GPT-5.6 Terra~\cite{GPT-5} then rewrites those clauses into a fluent instruction of two to four sentences under the prompt below, so that text-conditioned baselines receive input in the register they were designed for rather than a list of templated clauses.
The result is cached once per reference scene and reused thereafter.
Caching matters for fairness as much as for cost: every method that consumes text receives byte-identical input, and the judge is shown the same text as its layout criteria, so no method benefits from a more favorable phrasing.
Note that the specification is written once and frozen before any layout exists, so the judge scores every method against text that was fixed independently of the layouts it evaluates, avoiding circularity between the specification and the scores.
\styledfileinput[]{prompts/1_instruction_synthesis.txt}{}
Two of these constraints exist to keep the instruction usable rather than merely fluent. Forbidding absolute directions and room dimensions is what lets the same text be applied to a differently shaped room at all. Forbidding instance numbers matters because the judge later sees a rendering in which two dining chairs are indistinguishable; an instruction that named \texttt{dining\_chair\_2} would ask for something no image can verify, and would leak an instance identity that no method is given.
Both the instruction and iRecall refer to the same category-level set of
relations. Distinctions between individual instances of the same category are
consequently outside what we claim to verify, by construction of the metric
rather than as a side effect of how the instruction is phrased.
For the living-room scene shown below, the extractor emits $41$ facts, which
reduce to $28$ distinct category-level triplets; the first few are
\styledfileinput[]{prompts/2_relation_triplets_example.txt}{}
and the cached instruction reads
\styledfileinput[]{prompts/3_cached_instruction_example.txt}{}
The compression is visible: $41$ geometric relations become four sentences that name categories and orderings but no distances, so the instruction preserves which objects belong together and loses how far apart they stood.

\subsection{Baselines with the Exact Reference Layout}
\label{sec:supp_3d_baselines}
The instruction above is the only channel through which the text-conditioned baselines see the reference scene, and it discards distances by construction. 
To separate the contribution of our method from that of its richer input, we additionally give three baselines the exact metric layout of the reference scene.
Every variant arranges the target room's own furniture on the target floor plan $\mathcal{F}_T$, and is given the reference as oriented bounding boxes, namely each object's category, extent, center and facing direction in meters, which InstructScene receives in the form of a relation graph.
None of them receives the reference scene's point clouds, which our method encodes with Concerto~\cite{Concerto}.
The layout itself, however, contains every quantity that the reported metrics are computed from.

\noindent\textbf{LayoutVLM.}
LayoutVLM~\cite{LayoutVLM} is driven by a vision-language model that proposes spatial constraints, which its differentiable optimizer then solves.
We append the reference layout to its instruction as text, listing the reference floor polygon followed
by every reference object with its category, footprint, height, center and facing direction.
The model is asked to reproduce which objects form a group, their relative offsets and orientations and
their clearances to walls, as closely as the target room's shape allows.
The optimizer and all other settings are left unchanged. 
The appended reference block lists the same floor polygon and objects as the reference part of the GPT-6 Astra~\cite{Astra} request below.

\noindent\textbf{InstructScene.}
InstructScene~\cite{Instructscene} contains no large language model.
It first generates a semantic scene graph from the instruction, with object categories as nodes and spatial predicates as edges, and then generates the layout from that graph.
We therefore bypass the first stage and supply the graph directly.
The edges are computed from the reference layout with the same relation rules that InstructScene uses
to build its training graphs, so the graph is the ground-truth graph of the reference scene rather than a prediction from text.
The nodes are the target inventory: each target object is matched to an unused reference object of the same category with the closest extent and inherits its relations, while unmatched target objects enter as isolated nodes and unmatched reference objects are dropped.
As in the original model, each node also carries InstructScene's quantized OpenShape~\cite{OpenShape} code of its target asset.
The graph-to-layout stage then generates poses for the target objects, and their own assets and sizes are kept.
As in its original form, this stage is not conditioned on the floor plan.

\noindent\textbf{GPT-6 Astra.}
The general-purpose model receives a single request without tools, examples or layout heuristics.
The request opens with the task, which is to reproduce the functional arrangement of the reference room as closely as possible while keeping the layout collision-free, inside the floor polygon and clear of doors.
It then lists the reference floor polygon, its doors and windows, and every reference object with its category, extent, height, center and heading, followed by the cached instruction of the reference scene.
Next it lists the target floor polygon with its doors and windows, together with the target inventory, giving each object an identifier, a category and an extent.
In both rooms every door is listed with the $0.75\,\mathrm{m}$ clearance rectangle in front of it.
The model returns one position and heading per object, while extents and heights stay fixed, and it is required to place every inventory object exactly once.

The system prompt fixes only the coordinate convention and the output format:
\styledfileinput[]{prompts/4_astra_system_prompt.txt}{}
The user message follows the template below, where italic fields are filled from the two rooms:
\styledfileinput[]{prompts/5_astra_user_message_template.txt}{}
Like the other baselines, all three are evaluated on the same pruned target inventory as our method.

\subsection{Metrics and Judging Protocol}

\noindent\textbf{Instruction Recall (iRecall).}
iRecall~\cite{Instructscene} reapplies the relation extractor to the generated layout, measuring how much of the reference specification is preserved.
Matching is existential and at the level of object categories: a reference triplet counts as realised when some pair of objects of the corresponding categories stands in that relation in the output.
No instance correspondence is therefore required, which is what allows the metric to be applied unchanged to methods that produce none.
Categories are first normalized to functional groups, since 3D-FRONT~\cite{3D-Front} splits interchangeable furniture across several labels and a target room's loveseat would otherwise never match a reference room's multi-seat sofa even when it stands exactly where the sofa stood and serves the same role.
Only clear synonyms are merged, sofas with sofas and beds with beds.
A coffee table and a dining table stay distinct, because \emph{in front of the coffee table} and \emph{in front of the dining table} are different requirements.
Triplets are compared as sets, so a relation stated twice by two objects of the same category on the same side of a third counts once. Directional labels differ by side, so a group arranged around a shared anchor is represented by several distinct relations, each of which has to be realised independently.

\noindent\textbf{Judging Protocol.}
Collision-free and in-boundary rates follow LayoutVLM~\cite{LayoutVLM}'s geometric definitions, in-boundary using the same polygon-buffer semantics. The positional and rotational coherency ratings and the PSA score follow its judging protocol, including its rubric prompts verbatim and GPT-5.6 Terra~\cite{GPT-5} as the judge, with one deliberate change.
LayoutVLM judges against a user-written description in a setting where no reference scene exists.
Ours does exist, and as the example above shows the synthesized instruction is a lossy projection of it, so distances, spacing and grouping never reach the judge through text alone.
We therefore show the reference room's own rendering alongside the produced one, on top of the instruction, while the instruction remains the criterion being judged against.

\begin{figure*}[t]
    \centering
    \includegraphics[width=\linewidth]{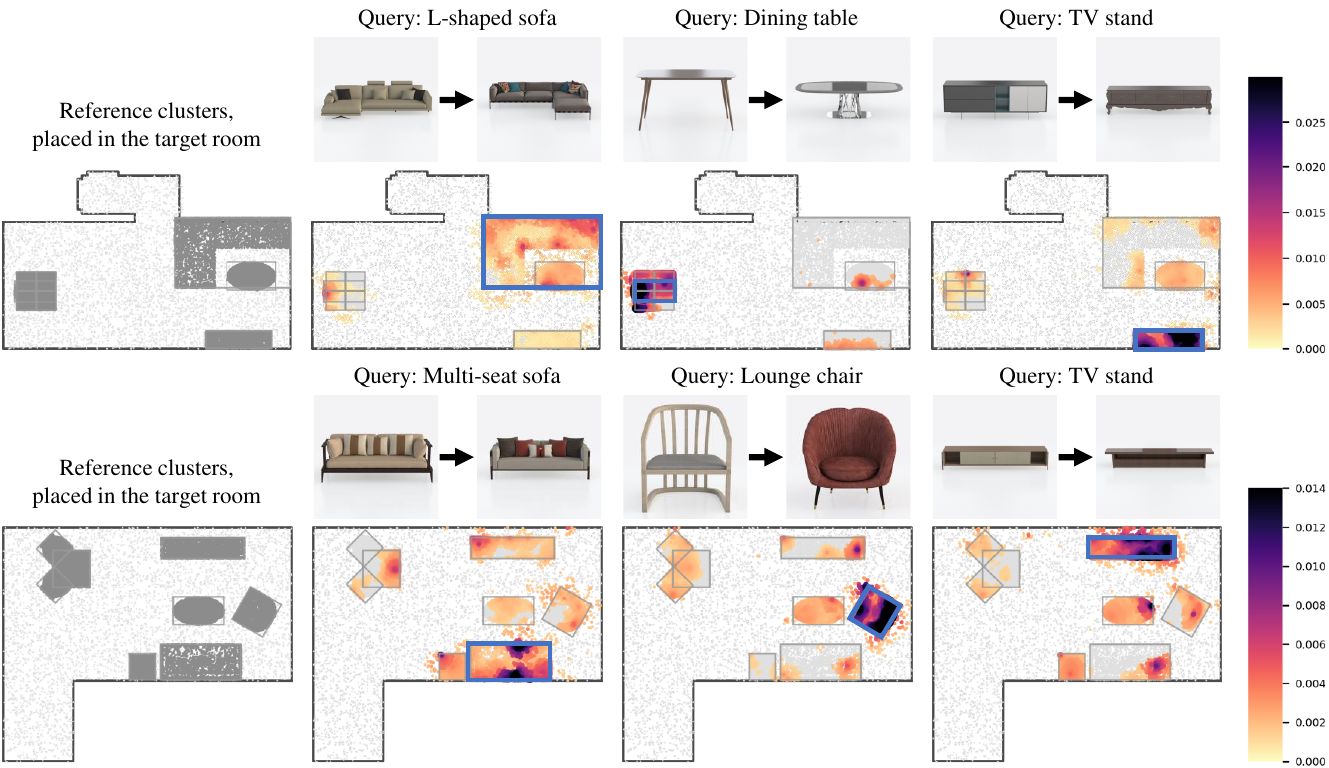}
    \caption{\textbf{Cross-Scene Attention Visualization.} Target object tokens selectively attend to their semantically corresponding reference furniture clusters, demonstrating robust semantic alignment under inventory and layout mismatches.}
    \label{fig:cross_attn}
\end{figure*}

\section{Additional Results}
\subsection{Analysis} 

\begin{table}[t]
\centering
\caption{\textbf{Effect of Cluster Granularity.} In boundary rate is measured before the projection step, and iRecall was measured for relational fidelity.}
\label{tab:cluster_granularity}
\small
\begin{tabular}{lccccc}
\toprule
Reference cluster size & 3 & 5 & 7 & 9 & 11 \\
\midrule
IB (before projection) $\uparrow$ & 93.7 & \textbf{93.9} & 91.7 & 88.6 & 87.9 \\
iRecall $\uparrow$ & 56.7 & 62.9 & \textbf{64.3} & 64.2 & 64.2 \\
\bottomrule
\end{tabular}
\end{table}
\noindent \textbf{Sensitivity Analysis on Cluster Granularity.} The number of objects per reference cluster controls the granularity of structural transfer, where extreme values fail for complementary reasons (\cref{tab:cluster_granularity}). 
In-boundary rates are reported prior to projection because the projection step forcibly clamps all objects into $\mathcal{F}_T$, driving containment near $100\%$ regardless of the initial layout. 
Pre-projection containment is thus a more informative metric because extensive clamping overwrites the optimized layout, sacrificing the relational structure established during refinement.
Small clusters preserve insufficient structure to transfer effectively. 
At three objects per cluster, iRecall drops to 56.7 because small clusters yield too few intra-cluster relations to be meaningfully preserved.
Beyond seven objects, additional intra-cluster structure yields no meaningful gain in preserved relations.
Conversely, containment exhibits the opposite trend. 
Larger clusters occupy larger footprints and are placed as rigid units, meaning a single misplacement forces all constituent objects outside $\mathcal{F}_T$. 
Consequently, the pre-projection in-boundary rate drops from $93.9\%$ at five objects to $87.9\%$ at eleven. 
In the limit, cluster decomposition loses its purpose because a cluster spanning most of the room reduces transfer to a rigid copy of the reference layout, which differing room boundaries render invalid. 
We therefore select a reference cluster size of seven, representing the minimal size required for relational fidelity to saturate before containment degrades significantly.

\noindent \textbf{Cross-Scene Attention at Inference.}
During inference, the stage 2 object decoder conditions on scene tokens derived from a transplanted reference layout, despite an object inventory mismatch between the reference and target rooms.
To examine what the decoder attends to under such mismatches, we inspect the stage 2 cross-attention at inference time.
Each query token's weights over the $768$ scene tokens, averaged across the decoder's six layers and eight heads, are projected onto the assembled scene point cloud, which comprises the placed reference clusters together with the target floor (\cref{fig:cross_attn}).
Cross-attention concentrates on the transferred reference furniture rather than on free space, aligning each query with the reference object that plays the same role even when the two assets differ in shape or category.
To confirm that identity rather than proximity drives this, we hold every position fixed and swap only object identities, meaning category, size, and OpenShape~\cite{OpenShape} features, between target objects.
Measured as total variation over the scene tokens, a slot's attention moves toward the map its newly assigned identity produces in its own slot in $890$ of $894$ trials.
An identity-preserving control, in which the paired objects hold the same asset at the same scale so the swapped token is numerically identical, leaves the map untouched.
The decoder therefore reads cross-scene context by object identity, and not by position alone.

\subsection{Relocating Instances to a New Floor Plan}
\label{sec:app_relocation}
\begin{figure*}[t]
    \centering
    \includegraphics[width=\linewidth]{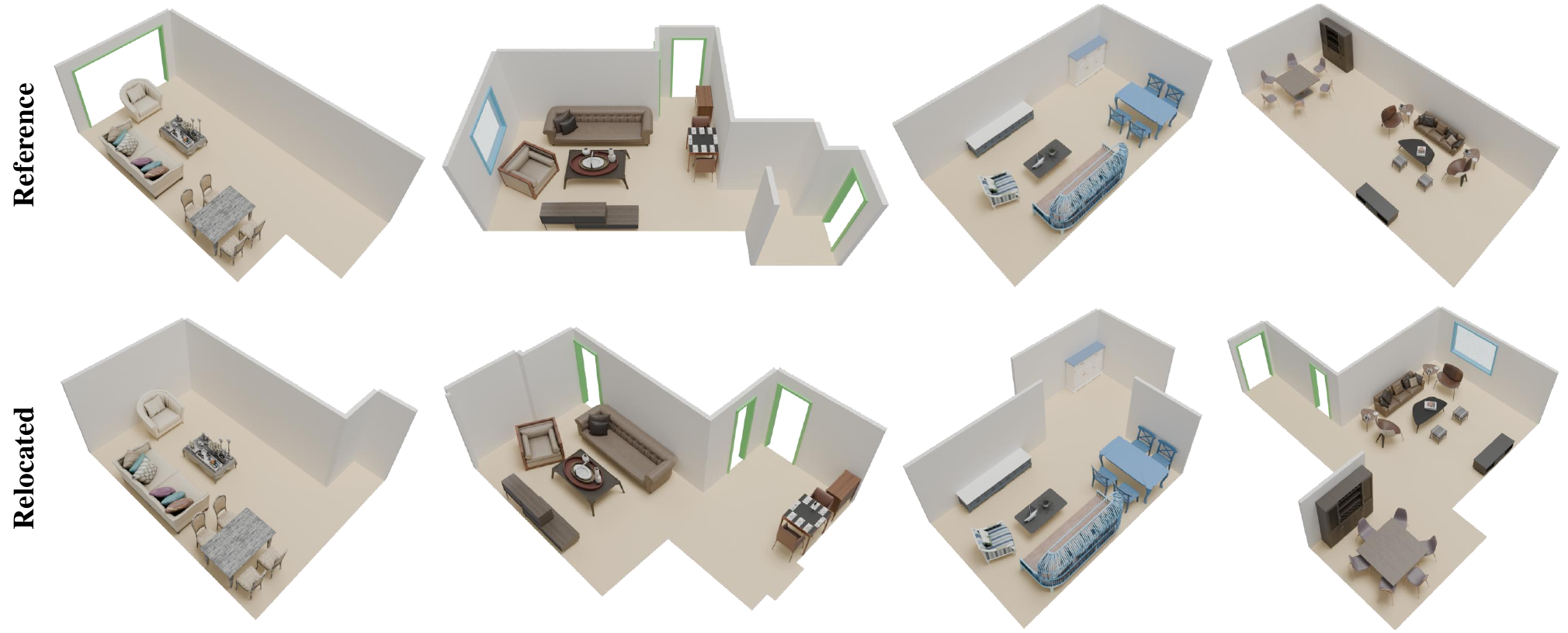}
    \caption{\textbf{Same-Inventory Relocation.} Paired reference and target room examples showing layout adaptation with identical object inventories across diverse room geometries.}
    \label{fig:relocation}
\end{figure*}
A natural special case of scene retargeting keeps the object inventory fixed and changes only the room, such as when a household moves its furniture to a home with a different floor plan. 
We instantiate this task as \emph{relocation}, where the reference scene supplies both the exemplar arrangement and the target inventory $\mathcal{O}_T=\{o^R_i\}_{i=1}^{N_R}$, while the target room contributes only its boundary $\mathcal{F}_T$ and openings $\mathcal{B}_T$. 
No component is retrained or tuned, and the exact same checkpoints and hyperparameters are used. 
Because every target object \emph{is} a reference object, the object-level decoder of stage~2 has nothing to infer. 
After stage~1, the placed clusters already constitute a complete layout in which every intra-cluster relation of the reference holds by construction, and the correspondence matrix $\mathbf{M}$ of stage~3 is the identity.
The framework therefore reduces to the two stages that reason about the new room. 
The placement network relocates each functional cluster onto $\mathcal{F}_T$, and the analogical refinement reestablishes the reference wall clearances. 
~\cref{fig:relocation} illustrates relocating reference rooms into distinct target floor plans of varying shapes, sizes, and opening layouts. 
Simply pasting the reference coordinates into these rooms would leave furniture outside the boundary or inside doorways wherever the plans differ, which is precisely what cluster placement and refinement repair. 
The result is a family of physically valid rooms that all keep the same original object inventory.

\subsection{Qualitative Results}
\begin{figure*}[t]
\centering
\includegraphics[width=0.9\linewidth]{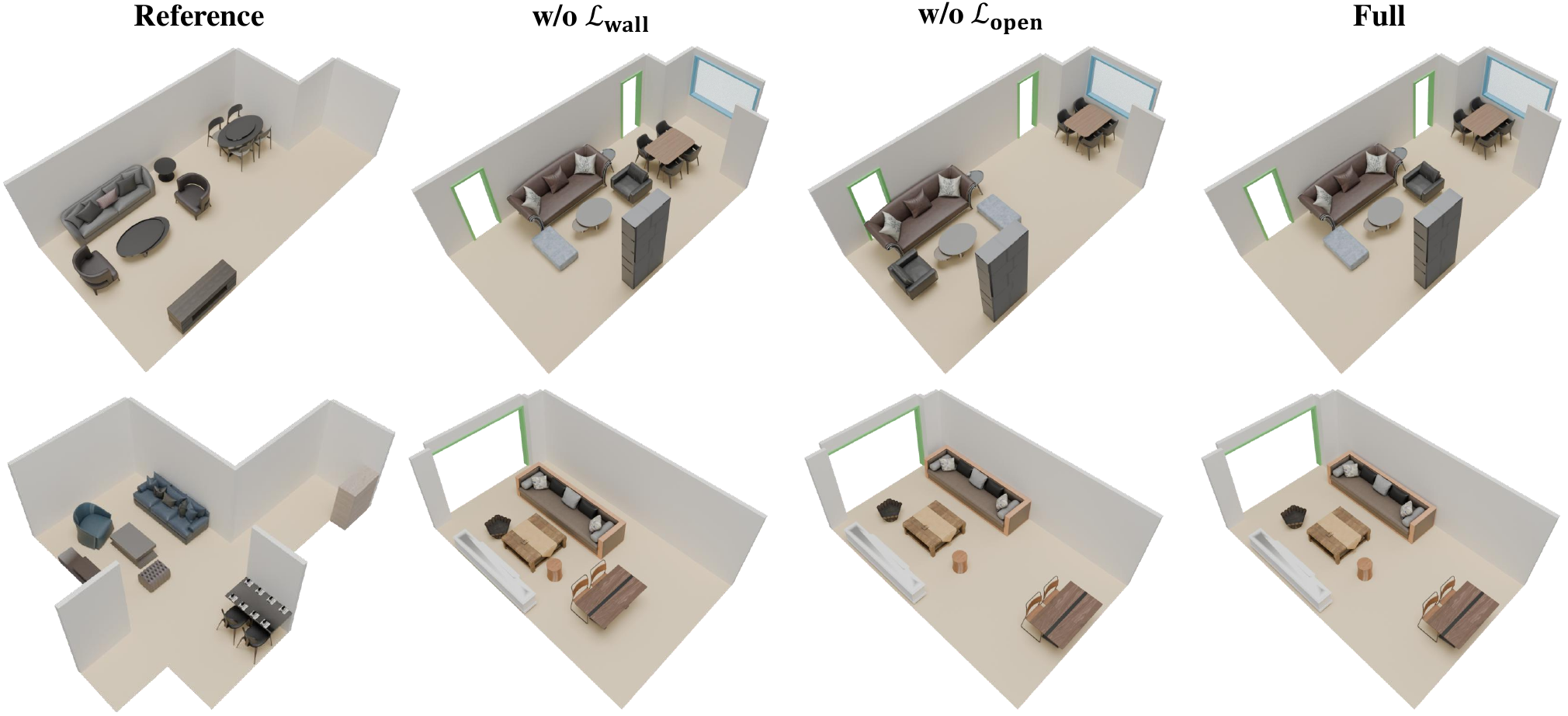}
\caption{\textbf{Effect of Wall Clearance and Opening Terms.} Omitting $\mathcal{L}_{\mathrm{wall}}$ alters relative wall clearance, whereas omitting $\mathcal{L}_{\mathrm{open}}$ obstructs room access. The full model succeeds on both fronts.}
\label{fig:loss_comparison}
\end{figure*}

\noindent\textbf{Effect of the Wall Clearance and Opening Refinement.} 
As shown in \cref{fig:loss_comparison}, omitting either spatial term degrades layout quality. 
Without $\mathcal{L}_{\mathrm{wall}}$, objects drift away from room boundaries, failing to preserve the relative object to wall clearance present in the reference scene. 
Removing $\mathcal{L}_{\mathrm{open}}$ results in objects blocking doorways and windows, compromising basic room accessibility. 
The full model maintains proper wall distance relationships while keeping openings unobstructed.

\begin{figure*}[t]
\centering
\includegraphics[width=0.9\linewidth]{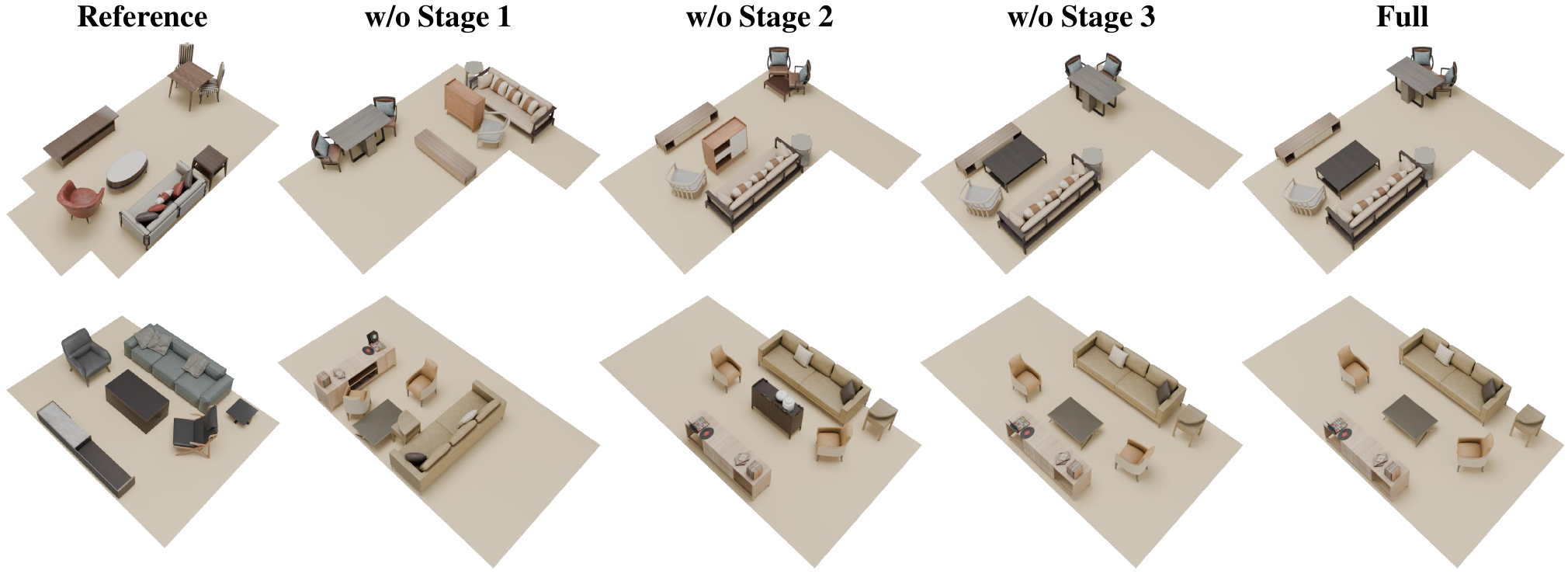}
\caption{\textbf{Visualization of Ablation Study.} 
Stage 1 keeps functional clusters intact within the target room, stage 2 assigns target objects to fitting positions under mismatched inventories, and stage 3 refines object relationships.
}

\label{fig:ablation}
\end{figure*}

\noindent\textbf{Ablation Studies.}
\cref{fig:ablation} visualizes the individual contributions of our core modules. 
Without stage 1 cluster placement, attempts to resolve physical collisions and boundary constraints force aggressive projection into the floor plan, which completely breaks the underlying layout structure. 
Without stage 2, target objects are assigned by explicit matching alone, which misplaces objects whose shape or count differs from the reference.
Finally, omitting stage 3 analogical refinement leaves fine object relationships unadjusted and fails to prune redundant assets.

\begin{figure*}[t]
\centering
\includegraphics[width=0.9\linewidth]{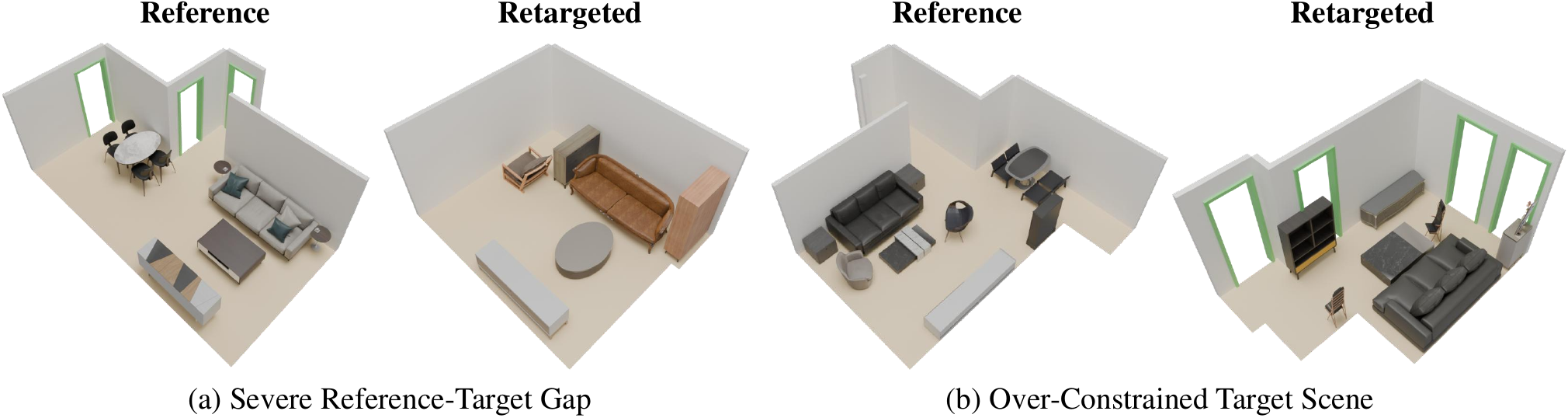}
\caption{\textbf{Failure Scenarios.} (a) Severe inventory mismatches between reference and target rooms lead to unnatural arrangements. (b) Highly constrained rooms cause conflicting spatial demands.}
\label{fig:failure_cases}
\end{figure*}

\noindent\textbf{Failure Cases.}
Our framework exhibits limitations in two challenging scenarios (\cref{fig:failure_cases}). 
First, a severe inventory gap between the reference and target room can cause awkward object groupings (\cref{fig:failure_cases} (a)). 
Second, in highly constrained target scenes, simultaneously satisfying opening clearance, wall proximity, and collision avoidance becomes geometrically unfeasible, leaving conflicting spatial demands unresolved (\cref{fig:failure_cases} (b)).

\begin{figure*}[t]
\centering
\includegraphics[width=\linewidth]{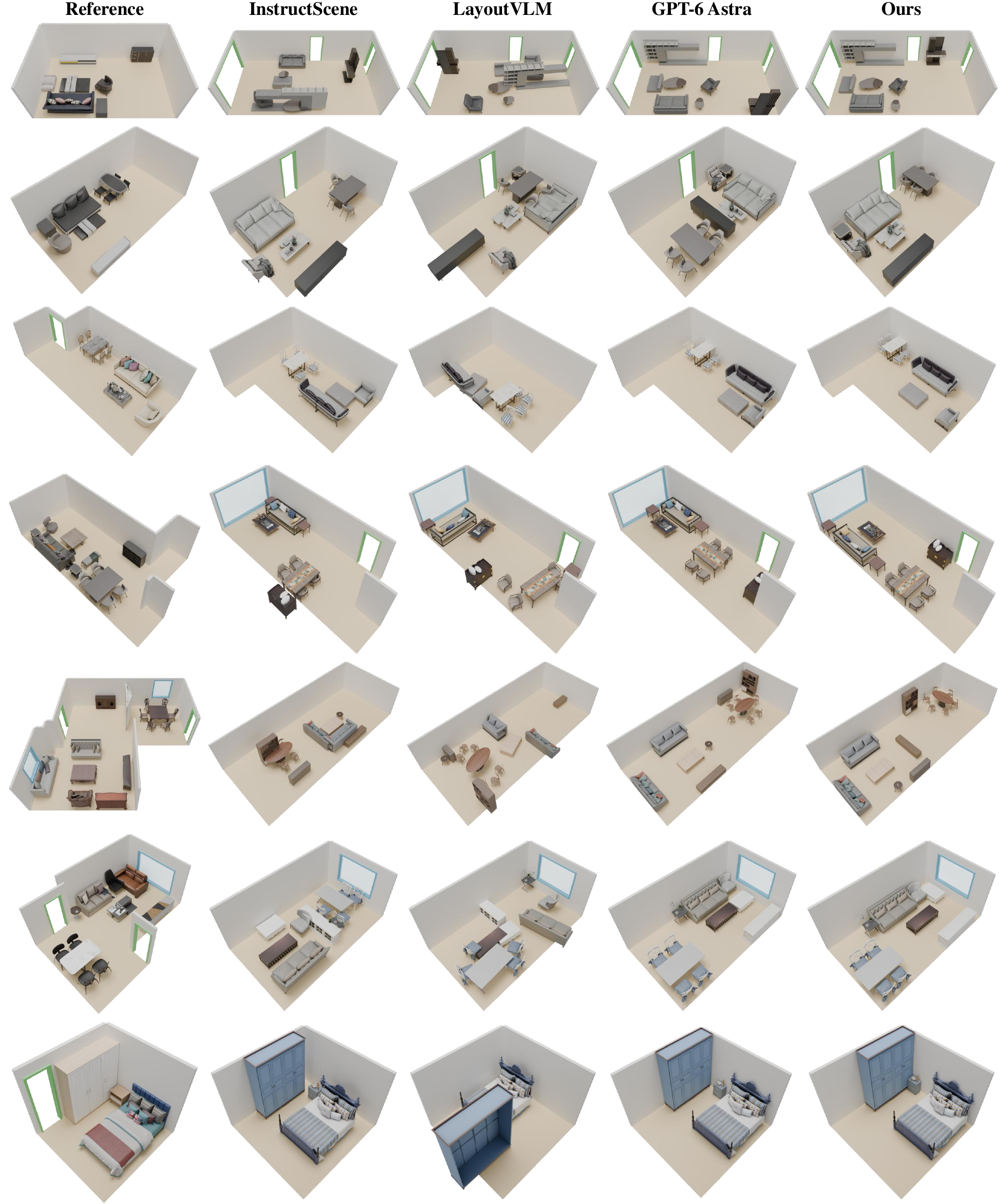}
\caption{\textbf{Additional Qualitative Comparisons.} Extended visual comparisons between baseline methods and our framework across diverse 3D-Front~\cite{3D-Front} scenes.} 
\label{fig:Additional Qualitative}
\end{figure*}

\noindent\textbf{Additional Qualitative Comparisons.}
\cref{fig:Additional Qualitative} provides additional qualitative retargeting results extending Fig. 4 of the main manuscript.
Baseline approaches such as InstructScene~\cite{Instructscene} and LayoutVLM~\cite{LayoutVLM} frequently suffer from fractured cluster arrangements, severe object overlaps, and wall collisions when adapting layouts to varying room shapes. 
GPT-6 Astra~\cite{Astra} produces physically valid layouts but loses finer details of the reference, such as relative offsets, orientations, and wall clearances.
In contrast, our framework faithfully maintains functional group structures and spatial relationships while adjusting overall placement to fit target floor plan boundaries without physical violations.

\newpage

\clearpage
\clearpage
\end{document}